\documentclass[10pt,journal,compsoc]{IEEEtran}

\usepackage[nocompress]{cite}
\ifCLASSINFOpdf
 
\else

\fi

\usepackage{bm}

\usepackage{amsmath}

\usepackage{mdwmath}

\usepackage{eqparbox}
\usepackage{epsfig}
\usepackage{graphicx}
\usepackage{amsmath}
\usepackage{amssymb}
\usepackage{pifont}
\usepackage{bbding}
\usepackage{url}            
\usepackage{booktabs}       
\usepackage{tabu}           
\usepackage{multirow}
\usepackage{graphicx}
\usepackage{algorithm}
\usepackage{algorithmicx}
\usepackage{algpseudocode}
\usepackage{amsmath,amssymb}
\usepackage{array}
\usepackage[dvipsnames]{xcolor}
\usepackage[pagebackref=true,breaklinks=true,letterpaper=true,colorlinks,bookmarks=false]{hyperref}

\usepackage{cite}

\usepackage{makecell}
\usepackage{tabularx}
\usepackage{multirow}
\usepackage{pifont}
\usepackage{multicol}
\usepackage{array}

\newcommand{\paragrapha}[2]{\vspace{#1}\noindent\textbf{#2}}

\usepackage{color}
\usepackage{adjustbox}
\usepackage{colortbl}
\usepackage{pifont}

\newcommand{\ie}{\textit{i}.\textit{e}.}
\newcommand{\etal}{\textit{et al}.}
\def\etc{\emph{etc}}
\def\eg{\emph{e.g}.}

\newcommand{\tabincell}[2]{\begin{tabular}{@{}#1@{}}#2\end{tabular}}  

\definecolor{Gray}{gray}{0.95}

\begin{document}
\title{AdaPoinTr: Diverse Point Cloud Completion with Adaptive Geometry-Aware Transformers}

\author{
        Xumin Yu,~\IEEEmembership{Student Member,~IEEE},
        Yongming Rao,~\IEEEmembership{Student Member,~IEEE},
        Ziyi Wang,~\IEEEmembership{Student Member,~IEEE},
        Jiwen Lu,~\IEEEmembership{Senior Member,~IEEE},
        Jie Zhou,~\IEEEmembership{Senior Member,~IEEE}
\IEEEcompsocitemizethanks{
\IEEEcompsocthanksitem The authors are with Beijing National Research Center for Information Science and Technology (BNRist), the State Key Lab of Intelligent Technologies and Systems, and the Department of Automation, Tsinghua University, Beijing, 100084, China. Email:~yuxm20@mails.tsinghua.edu.cn;~raoyongmimg95@gmail.com;~wziyi22@mails.tsinghua.edu.cn;~lujiwen@tsinghua.edu.cn;~jzhou@tsinghua.edu.cn.
}
}

\markboth{}%
{}
\IEEEtitleabstractindextext{
\begin{abstract}
Point clouds captured in real-world scenarios are often incomplete due to the limited sensor resolution, single viewpoint, and occlusion. Therefore, recovering the complete point clouds from partial ones becomes an indispensable task in many practical applications. In this paper, we present a new method that reformulates point cloud completion as a set-to-set translation problem and design a new model, called \textit{\textbf{PoinTr}}, which adopts a Transformer encoder-decoder architecture for point cloud completion. By representing the point cloud as a set of unordered groups of points with position embeddings, we convert the input data to a sequence of point proxies and employ the Transformers for generation. To facilitate Transformers to better leverage the inductive bias about 3D geometric structures of point clouds, we further devise a geometry-aware block that models the local geometric relationships explicitly. The migration of Transformers enables our model to better learn structural knowledge and preserve detailed information for point cloud completion. {\color{black} Taking a step towards more complicated and diverse situations, we further propose \textit{\textbf{AdaPoinTr}} by developing an adaptive query generation mechanism and designing a novel denoising task during completing a point cloud.  Coupling these two techniques enables us to train the model efficiently and effectively: we reduce training time (by 15x or more) and improve completion performance (over 20\%).} Additionally, we propose two more challenging benchmarks with more diverse incomplete point clouds that can better reflect real-world scenarios to promote future research. {\color{black} We also show our method can be extended to the scene-level point cloud completion scenario by designing a new geometry-enhanced semantic scene completion framework.} {\color{black} Extensive experiments on the existing and newly-proposed datasets demonstrate the effectiveness of our method, which attains 6.53 CD on PCN, 0.81 CD on ShapeNet-55 and 0.392 MMD on real-world KITTI, surpassing other work by a large margin and establishing new state-of-the-arts on various benchmarks. Most notably, AdaPoinTr can achieve such promising performance with higher throughputs and fewer FLOPs compared with the previous best methods in practice. The code and datasets are available at \url{https://github.com/yuxumin/PoinTr}. }
\end{abstract}

}

\maketitle

\IEEEdisplaynontitleabstractindextext
\IEEEpeerreviewmaketitle

\ifCLASSOPTIONcompsoc
\IEEEraisesectionheading{\section{Introduction}\label{sec:introduction}}
\else
\section{Introduction}
\label{sec:introduction}
\fi
\IEEEPARstart{R}{ecent} developments in 3D sensors largely boost researches in 3D computer vision. One of the most commonly used 3D data format is the point cloud, which requires less memory to store but convey detailed 3D shape information. However, point cloud data from existing 3D sensors are not always complete and satisfactory because of inevitable self-occlusion, light reflection, limited sensor resolution, \etc. Therefore, recovering complete point clouds from partial and sparse raw data becomes an indispensable task with ever-growing significance.

Over the years, researchers have tried many approaches to tackle this problem in the realm of deep learning. Early attempts on point cloud completion~\cite{DBLP:conf/cvpr/DaiQN17,DBLP:conf/iccv/HanLHKY17,DBLP:journals/corr/SharmaGF16,DBLP:conf/cvpr/StutzG18,DBLP:conf/cvpr/NguyenHTPY16,DBLP:conf/iros/VarleyDRRA17,DBLP:conf/nips/LiuTLH19,DBLP:conf/cvpr/LiuFXP19,yang20173d,wang2017shape} try to migrate mature methods from 2D completion tasks to 3D point clouds by voxelization and 3D convolutions. However, these methods suffer from a 
heavy computational cost that grows cubically as the spatial resolution increases. With the success of PointNet~\cite{DBLP:conf/cvpr/QiSMG17} and PointNet++~\cite{DBLP:conf/nips/QiYSG17}, directly processing 3D coordinates becomes the mainstream of point cloud based 3D analysis. The technique is further applied to many pioneer work~\cite{DBLP:conf/iclr/AchlioptasDMG18,PCN,TopNet,PFNet,DBLP:journals/corr/abs-1901-08906,groueix2018papier,sarmad2019rl} in point cloud completion task, in which an encoder-decoder based architecture is designed to generate complete point clouds. However, the bottleneck of such methods lies in the single feature vector produced in the encoding phase, where fine-grained information is lost and can hardly be recovered in the decoding phase.

Reconstructing complete point clouds is a challenging problem since the structural information required in the completion task runs counter to the unordered and unstructured nature of point cloud data. Therefore, learning structural features and long-range correlations among local parts of the point cloud becomes the key ingredient towards better point cloud completion. In this paper, we propose to adopt Transformers~\cite{Transformer}, one of the most successful architectures in Natural Language Processing (NLP), to learn the structural information of pairwise interactions and global correlations for point cloud completion. Our model, named \emph{PoinTr}, is characterized by five key components:  1) \emph{Encoder-Decoder Architecture}: We adopt the encoder-decoder architecture to convert point cloud completion as a set-to-set translation problem. The self-attention mechanism of Transformers models all pairwise interactions between elements in the encoder, while the decoder reasons about the missing elements based on the learnable pairwise interactions among features of the input point cloud and queries; 2) \emph{Point Proxy}: We represent the set of point clouds in a local region as a feature vector called \emph{Point Proxy}. The input point cloud is converted to a sequence of Point Proxies, which are used as the inputs of our Transformer model; 3) \emph{Geometry-aware Transformer Block}: To facilitate Transformers to better leverage the inductive bias about 3D geometric structures of point clouds, we design a geometry-aware block that models the geometric relations explicitly; 4) \emph{Query Generator}: Queries serve as the initial state of predicted proxies and we use dynamic queries instead of fixed queries in the decoder, which are generated by a query generation module that summarizes the features produced by the encoder and represents the initial sketch of the missing points; 5) \emph{Multi-Scale Point Cloud Generation}: We devise a multi-scale point generation module to recover the missing point cloud in a coarse-to-fine manner.

 \begin{figure}[t]
  \centering
  \includegraphics[width = \linewidth]{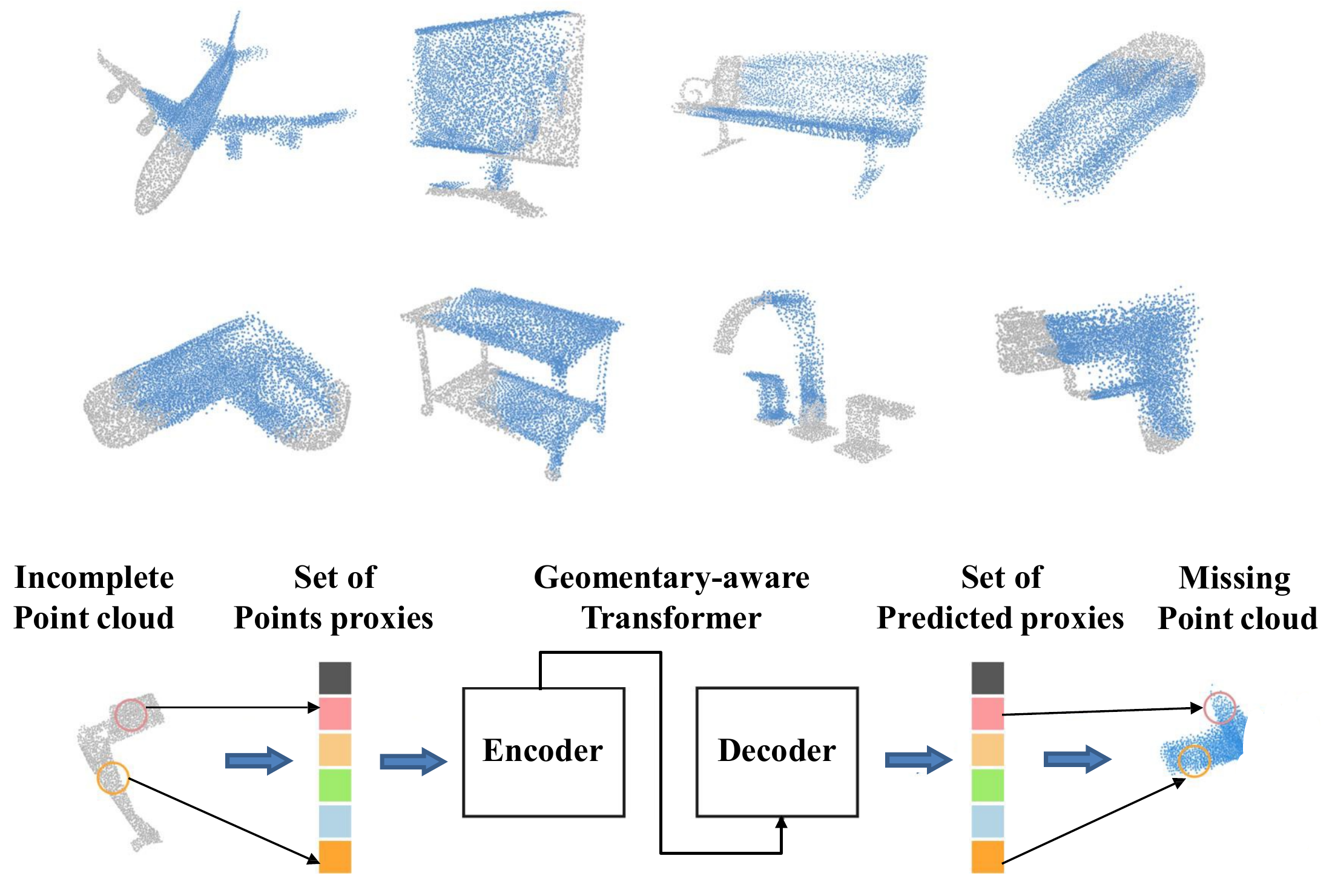}
  \caption{\small  \emph{PoinTr} is designed for point cloud completion task. It takes the downsampled partial point clouds as inputs ({\color{gray} gray points}), predicts the missing parts ({\color{blue} blue points}) and upsamples the known parts simultaneously. We propose to formulate the point cloud completion task as a set-to-set translation task and use a Transformer encoder-decoder architecture to learn the complex dependencies among the point groups. 
  }
  \label{fig:insight}
\end{figure}

{\color{black} 
Based on the proposed architecture, 
point cloud completion is reformulated as a set-to-set translation problem, as shown in Fig.~\ref{fig:insight}.
Given the incomplete point cloud ({\color{gray} gray points}) as the known part, we translate it into the unknown part ({\color{blue} blue points}) with our PoinTr, which enables us to fully exploit the interactions between known and unknown sets. Then a simple concatenation operation is performed to combine these two partial point clouds into a complete one. However, since these two parts are considered separately during the completion process, the concatenation operation in the $\mathcal{R}^3$ space brings the issue that the final prediction may be discontinuous in appearance. Therefore, we further propose \textit{\textbf{AdaPoinTr}} based on PoinTr, which combines the known and unknown parts by concatenating corresponding queries in the embedding space $\mathcal{R}^Q$ rather than in the $\mathcal{R}^3$ space (See $\S$~\ref{comp3} for detailed explanation). In this way, the known and unknown parts can be considered jointly during the completion process in a unified manner. Technically, we develop an adaptive query generation mechanism to deal with diverse situations. For example, a car from the LiDAR-scan may miss over 90\% points while another one may miss only 50\% points. 
Moreover, an auxiliary denoising task is designed to effectively make the optimization more stable and efficient, as it alleviates the problem caused by low-quality queries at the very beginning of the training. As shown in Fig.~\ref{fig:improvement}, AdaPoinTr can achieve better performance with only 8\% training epochs compared with PoinTr. Meanwhile, AdaPoinTr can produce more reasonable completion results without appearance issues even in challenging real-world situations.

\begin{figure}[t]
\centering
\includegraphics[width = \linewidth]{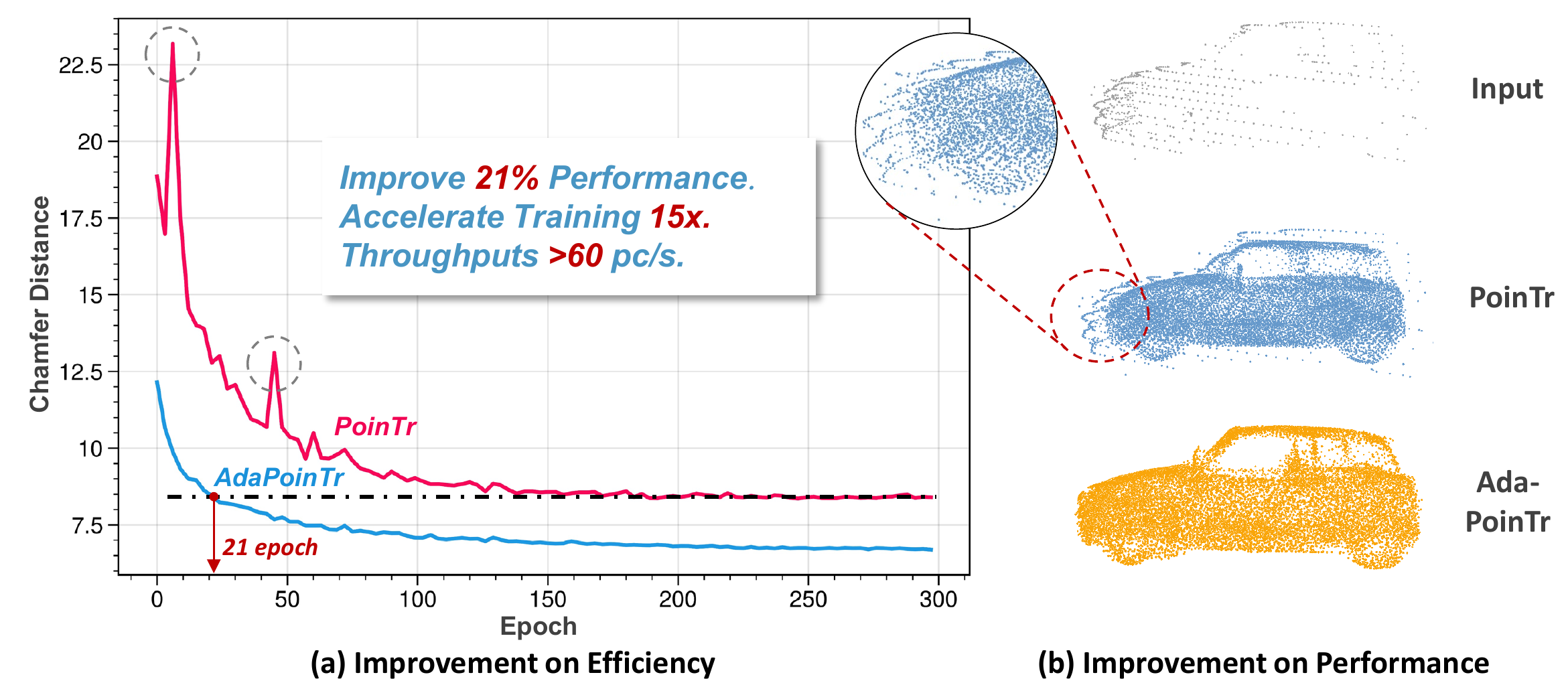}
\caption{\small We further propose AdaPoinTr by developing an adaptive query generation mechanism and designing a novel denosing task. We achieve improvement on both efficiency and performance with less training epochs, excellent throughputs and more stable training curve on wide-used PCN dataset.}
\label{fig:improvement}
\end{figure}
}

As another contribution, we argue that existing benchmarks are not representative enough to cover real-world scenarios of incomplete point clouds. Therefore, we introduce two more challenging benchmarks, ShapeNet-55 and ShapeNet-34, in short \emph{SN-55/34}, which contain more diverse tasks (\ie, joint upsampling and completion of point cloud), more object categories (\ie, from 8 categories to 55 categories), more diverse views points (\ie, from 8 viewpoints to all possible viewpoints) and more diverse level of incompleteness (\ie, missing 25\% to 75\% points of the ground-truth point clouds). {\color{black} In \emph{SN-55/34}, we adopt an online-cropping method to generate diverse incomplete point clouds, which is more flexible and efficient than other offline back-projecting benchmarks like PCN~\cite{PCN}, Completion3D~\cite{TopNet} and MVP~\cite{vrc}. 
Even though their samples are more realistic than those in our efficient \emph{SN-55/34}, they still did not take noises into consideration. At this point, we propose noised back-projecting by adding a scaled random noise to the depth map before generating incomplete point clouds. We change the online cropping method with this noised back-projecting method and obtain two new variants: Projected-ShapeNet-55 and Projected-ShapeNet-34. We show some samples of incomplete point cloud generation with different methods (blue columns) in Fig.~\ref{fig:dataset}. The noised back-projecting makes the completion task more difficult and avoids the trivial solution of simply combining the incomplete point clouds from different views, which are shown in the last two columns in the figure.

\begin{figure}[t]
\centering
\includegraphics[width = \linewidth]{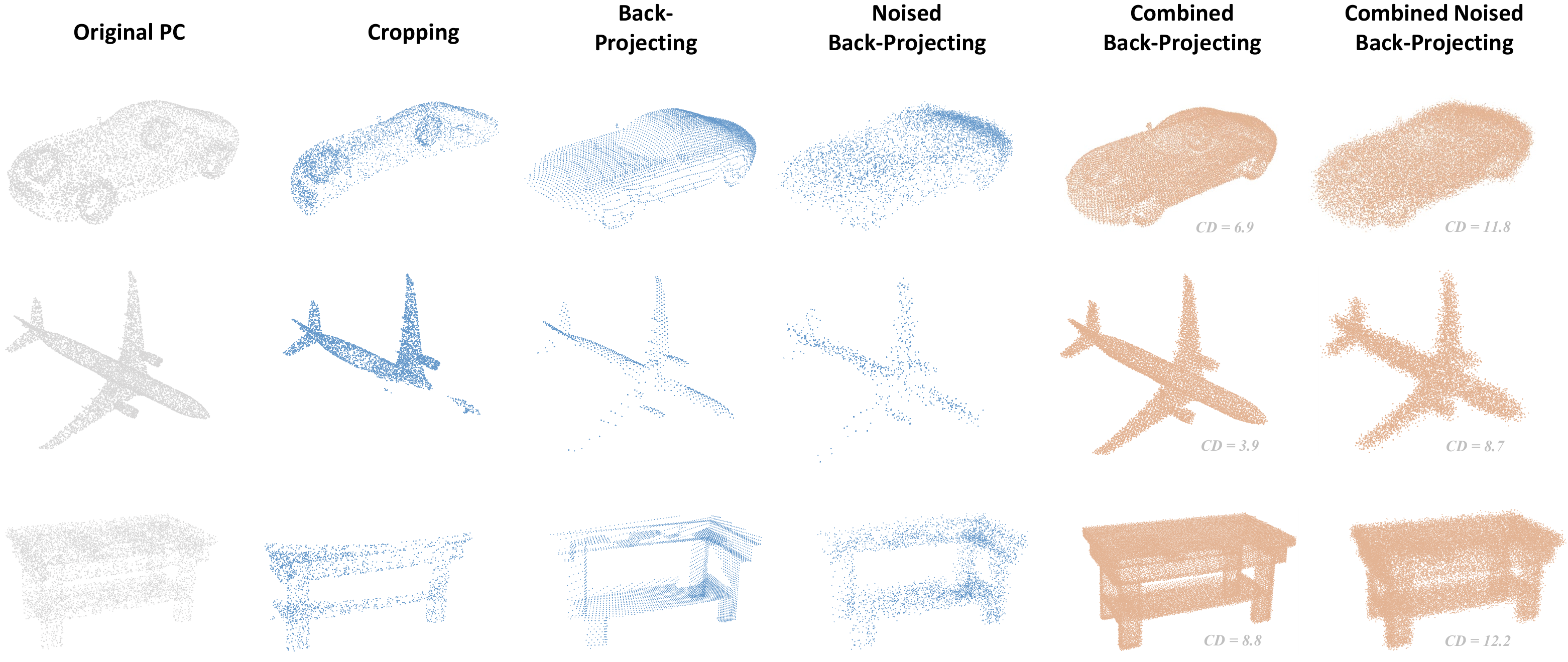}
\caption{\small There are three main-stream methods to generate incomplete point clouds from 3D objects, cropping, back-projecting, and noised back-projecting. We propose ShapeNet-55/34 and Projected-ShapeNet-55/34, which generate incomplete point clouds by cropping and noised back-projecting, respectively. Online cropping is more flexible and efficient, while noised back-projecting is a better approximation to real scans.}
\label{fig:dataset}
\end{figure}

Besides, we also explore the application potential of our proposed architecture in scene-level completion task. Due to the large scale of a point cloud scene, scene-level completion is always formulated as a voxel-based completion task, predicting the occupancy for the given voxel, which always lacks of considering detailed geometric information. Luckily, our model demonstrates its outstanding performance on learning point-wise structural features and building long-range correlations among local regions. Therefore, we introduce the proposed geometry-aware Transformer architecture to the scene-level completion task. Specifically, we propose a geometry-enhanced semantic scene completion framework, which can supplement voxel-based models with fine-grained geometric information, thus improving the completion performance on scenes. 

{\color{black} We conduct experiments on both object point cloud completion and semantic scene completion. For object point cloud completion, we evaluate our method on the newly proposed benchmarks, the widely used dataset PCN~\cite{PCN}, and the LiDAR-based dataset KITTI~\cite{KITTI}. Our AdaPoinTr establishes new state-of-the-art on all the mentioned benchmarks. Additionally, we evaluated our model on the MVP competition benchmark~\cite{mvpcompetition} and won the first place of the MVP Completion challenges in the ICCV 2021 Workshop~\cite{pan2021multi}. For scene-level point cloud completion, we follow the standard protocol of the semantic scene completion task~\cite{scc-survey} and evaluate our method on NYU Depth V2~\cite{silberman2012indoor} and NYUCAD~\cite{firman2016structured}.}} Extensive experiments demonstrate the effectiveness of our method in various application scenarios.  

This paper is an extended version of our conference paper~\cite{yu2021pointr}. We make several new contributions:
1) We propose an improved version of original PoinTr, AdaPoinTr, by adopting an adaptive query generation mechanism and designing a novel auxiliary denoising task during completion, which solves the appearance issue and enables us to establish new state-of-the-arts with less training times and satisfactory throughputs in practice.
2) we extend our method to the scene-level point cloud completion by designing a geometry-enhanced semantic scene completion framework and performing point-to-voxel translation with PoinTr; 
3) we present Projected-ShapeNet-55/34 dataset by generating incomplete samples with proposed noised back-projecting to better approximate the real scans. We also provide more in-depth analysis and visualization of our method.

\section{Related Work}

 \begin{figure*}[th]
  \centering
  \includegraphics[width = 1\linewidth]{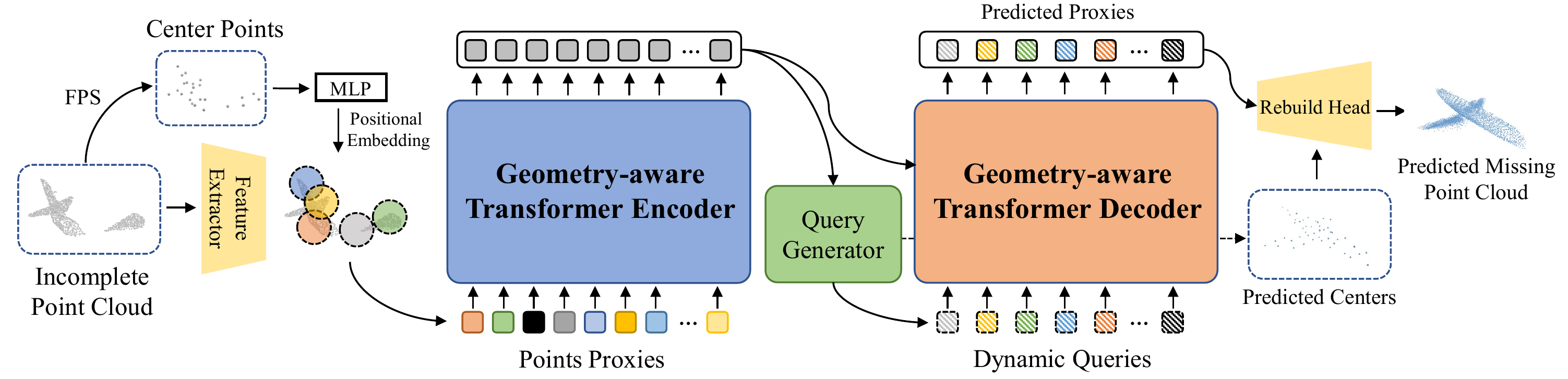}
  \caption{\small The Pipeline of \emph{PoinTr}. We first downsample the input partial point cloud to obtain the center points. Then, we use a lightweight DGCNN~\cite{wang2019DGCNN} to extract the local features around the center points. After adding the position embedding to the local feature, we use a Transformer encoder-decoder architecture to predict the point proxies for the missing parts. A simple MLP and a Rebuild Head are used to complete the point cloud based on the predicted point proxies in a coarse-to-fine manner.
  }
  \label{fig:pipeline}
\end{figure*}

\label{sec:relatedwork}
\paragrapha{0pt}{3D Shape Completion. } 
Traditional methods for 3D shape completion tasks often adopt voxel grids or distance fields to describe 3D objects~\cite{DBLP:conf/cvpr/DaiQN17,DBLP:conf/iccv/HanLHKY17,DBLP:conf/cvpr/StutzG18}. Based on such structured 3D representations, the powerful 3D convolutions are used and achieve a great success in the tasks of 3D reconstruction~\cite{DBLP:conf/eccv/ChoyXGCS16,DBLP:conf/eccv/GirdharFRG16} and shape completion~\cite{DBLP:conf/cvpr/DaiQN17,DBLP:conf/iccv/HanLHKY17,ShapeNet}. However, this group of methods~\cite{DBLP:conf/cvpr/SuJSMK0K18,DBLP:journals/tog/WangLGST17,DBLP:conf/cvpr/GrahamEM18} suffers from heavy memory consumption and computational burden. 
Different from the above methods, researchers gradually start to use unstructured point clouds as the representation of 3D objects, given the small memory consumption and strong ability to represent fine-grained details. Nevertheless, the migration from structured 3D data understanding to point cloud analysis is non-trivial, since the commonly used convolution operator is no longer suitable for unordered points clouds. PointNet and its variants~\cite{DBLP:conf/cvpr/QiSMG17,DBLP:conf/nips/QiYSG17} are the pioneering work to directly process 3D coordinates and inspire the researches in many downstream tasks. In the realm of point cloud completion, PCN~\cite{PCN} is the first learning-based architecture, which proposes an Encoder-Decoder framework and adopts a FoldingNet to map the 2D points onto a 3D surface by mimicking the deformation of a 2D plane. After PCN, many other methods~\cite{TopNet,PFNet,GRNet,liu2020morphing,yu2021pointr} spring up, pursuing point clouds completion in higher resolution with better robustness. {\color{black} VRCNet~\cite{vrc} proposes a variational framework with probabilistic modeling and relational enhancement, which effectively exploits 3D structural relations to predict complete shapes. }{\color{black} SnowflakeNet~\cite{xiang2021snowflakenet} proposes to model the generation of point clouds as a snowflake-like growth based on certain points in 3D space, where the point clouds are generated progressively from some parent points. And recently, LAKeNet~\cite{tang2022lake} proposes to consider predicting structured and topological information of 3D shape, which follow a keypoints-skeleton-shape prediction manner. SeedFormer~\cite{zhou2022seedformer} introduces a new shape representation, Patch Seeds, for point clouds and devises a novel Upsample Transformer during the generation.

}

{\color{black}
\paragrapha{5pt}{Semantic Scene Completion. } 
Semantic Scene Completion is an important task in 3D scene understanding, which aims to predict volumetric occupancy and semantic labels simultaneously from a single-view depth map or RGB-D image. For a long time, many methods consider these two tasks separately. SSCNet~\cite{song2017semantic} is the first to introduce Semantic Scene Completion by combining scene completion and semantic segmentation in an end-to-end way, proving that these two tasks can promote each other in the meantime. ESSCNet~\cite{zhang2018efficient} proposes Spatial Group Convolution (SGC) to group the input volume and then utilizes 3D sparse convolution for feature extraction. VVNet~\cite{guo2018view} further proposes to bridge 2D domain and 3D domain through a differentiable projection layer, which efficiently reduces the computational cost and makes it possible to extract the feature from multi-channel inputs. ForkNet~\cite{wang2019forknet} proposes to build a multi-branch architecture and alleviates problems caused by the limited training samples of real scenes by adopting generative models. CCPNet~\cite{zhang2019cascaded} argues to progressively restore the details of objects by adopting a cascaded context pyramid model, which improves the labeling coherence. Then some methods~\cite{garbade2019two, liu2018see} try to seek a way to leverage the complementary information of depth map, like RGB images. 3D CNN is employed to fuse two-stream inputs. DDRNet~\cite{li2019rgbd} proposes a light-weight dimensional decomposition residual block to fuse multi-scale RGB-D features. AICNet~\cite{li2020anisotropic} modifies the standard 3D convolution so that the kernels can be of various sizes. Sketch~\cite{chen20203d} proposes to guide the semantic prediction with an explicitly encoded 3D sketch-aware feature embedding, which contains rich geometric information. Recently, SISNet~\cite{cai2021semantic} exploits the relations between instance completion and scene completion in an interactive manner. They iteratively reconstruct instances within the scene and complete the entire scene, which effectively recovers the geometric patterns and semantic information compared with previous end-to-end methods. }

\paragrapha{5pt}{Transformers. } Transformers~\cite{Transformer} are first introduced as an attention-based framework in Natural Language Processing (NLP). Transformer models often utilize the encoder-decoder architecture and are characterized by both self-attention and cross-attention mechanisms. Transformer models have proven to be very helpful to tasks that involve long sequences thanks to the self-attention mechanism. The cross-attention mechanism in the decoder exploits the encoder information to learn the attention map of query features, which makes Transformers powerful in generation tasks. By taking advantage of both self-attention and cross-attention mechanisms, Transformers have a strong capability to handle long sequence input and enhance information communications between the encoder and the decoder. In the past few years, Transformers have dominated the tasks that take long sequences as input and gradually replaced RNNs~\cite{vinyals2015order} in many domains. Now they begin their journey in computer vision~\cite{DBLP:conf/naacl/DevlinCLT19,DBLP:journals/corr/abs-1905-03072,2019arXiv191108460S,DBLP:journals/corr/abs-1802-05751,yu2022point}. { ViT~\cite{dosovitskiy2020image} introduces Transformers into 2D domains, and DeiT~\cite{touvron2021training} explores a data-efficient training strategy for Vision Transformers}. While most efforts focus on learning vision Transformers on 2D data, the  applications on point clouds remain limited. Some preliminary explorations on recognition task have been implemented\cite{zhao2021point,guo2021pct}, while we introduce this architecture into 3d generation tasks like point cloud completion.

\section{Point Cloud Completion}

In this section, we will introduce the details of \emph{PoinTr}. We first elaborate the five key components of PoinTr in Section \ref{comp1}-\ref{comp2}. Then, we present the new adaptive query generation mechanism and auxiliary denoise task in Section~\ref{comp3}. Lastly, we show the learning objectives in Section~\ref{loss}.   The overall framework of our method is illustrated in Fig.~\ref{fig:pipeline}.
\label{sec:method}

\subsection{Set-to-Set Translation with Transformers}
\label{comp1}
The primary goal of our method is to leverage the impressive sequence-to-sequence generation ability of Transformer architecture for point cloud completion tasks. We propose to first convert the point cloud to a set of feature vectors, \textit{point proxies}, that represent the local regions in the point clouds (we will describe in Section~\ref{sec:proxy}). By analogy to the language translation pipeline, we model point cloud completion as a set-to-set translation task, where the Transformers take the point proxies of the partial point clouds as the inputs and produce the point proxies of the missing parts. Specifically, given the set of point proxies $\mathcal{F} = \{F_1, F_2, ..., F_N\}$ that represents the partial point cloud, we model the process of point cloud completion as a set-to-set translation problem:
\begin{eqnarray} \small
    \mathcal{V} = \mathcal{M}_E(\mathcal{F}), \quad \mathcal{H} = \mathcal{M}_D( \mathcal{Q}, \mathcal{V}),
\end{eqnarray}
where $\mathcal{M}_E$ and $\mathcal{M}_D$ are the encoder and decoder models, $\mathcal{V} = \{V_1, V_2, ..., V_N\}$ are the output features of the encoder,  $\mathcal{Q} =  \{Q_1, Q_2, ..., Q_M\}$ are the dynamic queries for the decoder, $\mathcal{H} =  \{H_1, H_2, ..., H_M\}$ are the predicted point proxies of the missing point cloud, and $M$ is the number of the predicted point proxies. The recent success in NLP tasks like text translation and question answering~\cite{devlin2018bert} have clearly demonstrated the effectiveness of Transformers to solve this kind of problem. Therefore, we propose to adopt a Transformer-based encoder-decoder architecture to solve the point cloud completion problem. 

The encoder-decoder architecture consists of $L_E$ and $L_D$ multi-head self-attention layers~\cite{Transformer} in the encoder and decoder, respectively. The self-attention layer in the encoder first updates proxy features with both long-range and short-range information. Then a feed-forward network (FFN) further updates the proxy features with an MLP architecture. The decoder utilizes self-attention and cross-attention mechanisms to learn structural knowledge. The self-attention layer enhances the local features with global information, while the cross-attention layer explores the relationship between queries and outputs of the encoder. To predict the point proxies of the missing parts, we propose to use dynamic query embeddings, {\color{black} which is different from the learnable static query embedding in ~\cite{carion2020end, zhang2022dino} (as shown in Fig.~\ref{fig:query} (a, b)).} This dynamic query scheme makes our decoder more flexible and adjustable for different types of objects and their missing information. More details about the Transformer architecture can be found in the supplementary material and~\cite{devlin2018bert,Transformer}.

Note that benefiting from the self-attention mechanism in Transformers, the features learned by the Transformer network are invariant to the order of point proxies, which is also the basis of using Transformers to process point clouds. Considering the strong ability to capture data relationships, we expect the Transformer architecture to be a promising alternative for deep learning on point clouds. 

\subsection{Point Proxy} \label{sec:proxy}

The Transformers in NLP take as input a 1D sequence of word  embeddings~\cite{Transformer}. To make 3D point clouds suitable for Transformers, the first step is to convert the point cloud to a sequence of vectors. A trivial solution is directly feeding the sequence of $xyz$ coordinates to the Transformers. However, since the computational complexity of the Transformers is quadratic to the sequence length, this solution will lead to an unacceptable cost. Therefore, we propose to represent the original point cloud as a set of \textit{point proxies}. A point proxy represents a local region of the point clouds. Inspired by the set abstraction operation in~\cite{DBLP:conf/nips/QiYSG17}, we first conduct \textit{furthest point sample (FPS)} to locate a fixed number $N$ of point centers $\{p_1, p_2, ..., p_N\}$ in the partial point cloud. Then, we use a lightweight DGCNN~\cite{wang2019DGCNN} with hierarchical downsampling to extract the feature of the point centers from the input point cloud. The point proxy $F_i$ is a feature vector that captures the local structure around $p_i$, which can be computed as:
\begin{equation} \small
    F_i = F'_i + \varphi(p_i), \label{eq:proxy}
\end{equation}
where $F'_i$ is the feature of point $p_i$ that is extracted using the DGCNN model, and $\varphi$ is another MLP to capture the location information of the point proxy. The first term represents the semantic patterns of the local region, and the second term is inspired by the position embedding~\cite{bello2019attention} operation in Transformers, which explicitly encodes the global location of the point proxy. The detailed architecture of the feature extraction model can be found in Supplementary Material.

\begin{figure}[t]
  \centering
  \includegraphics[width = \linewidth]{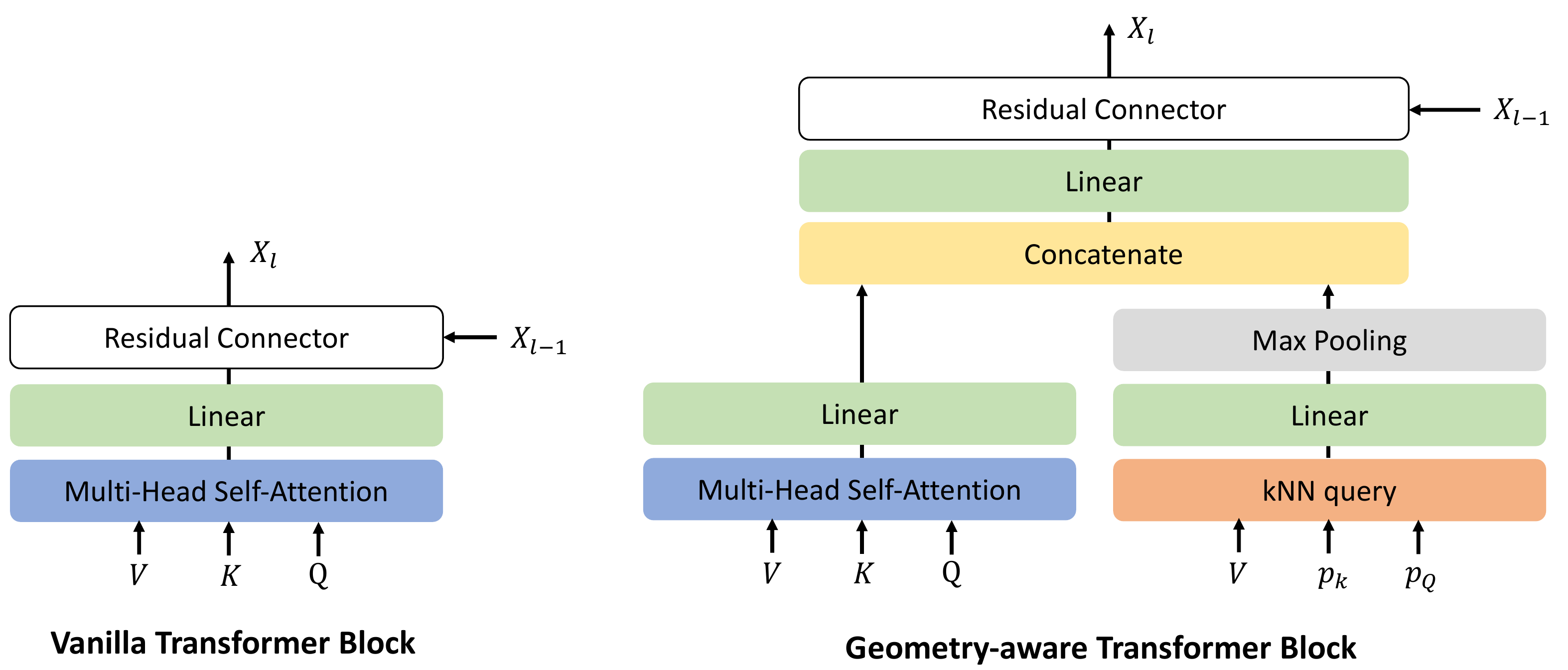}
  \caption{\small Comparisons of the vanilla Transformer block and the proposed geometry-aware Transformer block.}
  \label{fig:geo}
\end{figure}

\subsection{Geometry-aware Transformer Block} 

One of the key challenges of applying Transformers for vision tasks is that the self-attention mechanism in Transformers lacks some inductive biases inhered in conventional vision models like CNNs, which explicitly model the structures of vision data. To facilitate Transformers to better leverage the inductive bias about 3D geometric structures of point clouds, we design a geometry-aware block that models the geometric relations, which can be a plug-and-play module to incorporate with the attention blocks in any Transformer architecture. The details of the proposed block are shown in Fig.~\ref{fig:geo}. Different from the self-attention module that uses the feature similarity to capture the semantic relation, we propose to use kNN to capture the geometric relations in the point cloud:
\begin{equation} \small
    \phi(V_i) = \mathcal{M}(\text{Linear}([V_i, V_j - V_i])), \forall j: p^j_k \in \kappa(p_Q^i) , \label{eq:knnpath}
\end{equation}
where $\mathcal{M}$ is max-pooling operation, $V$ is the input features. Given the query coordinates $p_Q$, we gather the features whose coordinates $p_k$ locates within the k-neighborhood of $p_Q$, represented as $\kappa(p_Q)$. Then we follow the practice of DGCNN~\cite{wang2019DGCNN} to learn the local geometric structures via feature aggregation with a linear layer followed by the max-pooling operation as shown in Equ.~\ref{eq:knnpath}. The feature captured by kNN and the feature captured by multi-head self-attention are then concatenated and mapped to the original dimensions to form the output.

 \begin{figure*}[th]
  \centering
  \includegraphics[width = \linewidth]{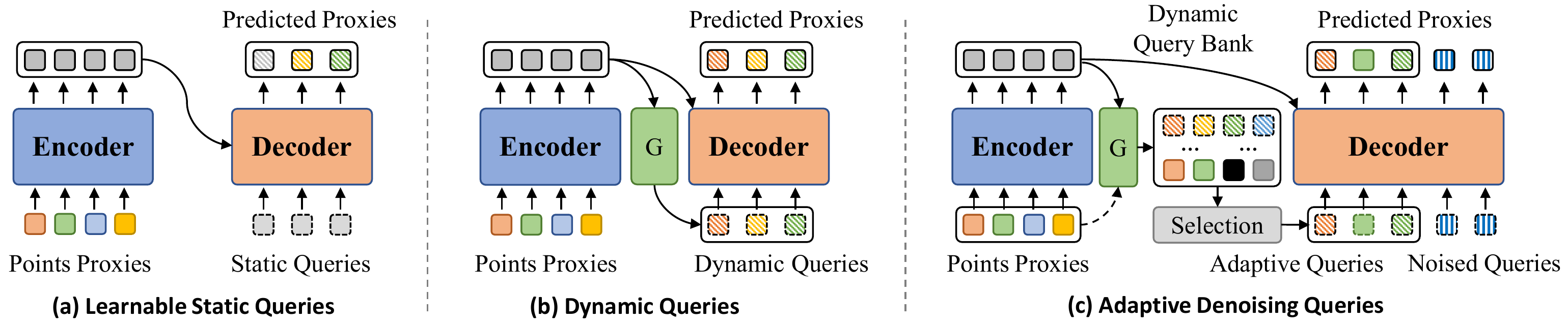}
  \caption{\small Three types of queries for transformer decoder. (a) Learnable static queries, which are usually used in DETR-style framework. (b) Dynamic queries proposed in PoinTr, which are generated based on the output features of the encoder. (c) Proposed adaptive denoising queries, which are selected from a dynamic query bank.
  }
  \label{fig:query}
\end{figure*}

\subsection{Query Generator}

The queries $\mathcal{Q}$ serve as the initial state of the predicted proxies. To make sure the queries correctly reflect the sketch of the complete point cloud, we propose a query generator module to generate the query embeddings dynamically conditioned on the encoder outputs $\mathcal{V}$. Specifically, we first summarize $\mathcal{V}$ with a linear projection to higher dimensions followed by the max-pooling operation. Similar to~\cite{PCN}, we use a linear projection layer to directly generate $M\times3$ dimension features that can be reshaped as the $M$ coordinates $\mathcal{C} = \{c_1, c_2, ..., c_M\}$. Lastly, we concatenate the global feature of the encoder and the coordinates, and use an MLP to produce the query embeddings $\mathcal{Q} =  \{Q_1, Q_2, ..., Q_M\}$, which can be formulated as follows:
\begin{equation} \small
    \mathcal{C} = \mathcal{P}(\mathcal{M}(\text{Linear}(\mathcal{V})), \quad 
    \mathcal{Q} = \text{MLP}([\mathcal{C}, \mathcal{M}(\text{Linear}(\mathcal{V})]),
\end{equation}
where $\mathcal{M}$ and $\mathcal{P}$ are max-pooling operation and coordinates projection, respectively.

\subsection{Multi-Scale Point Cloud Generation}
\label{comp2}

The goal of our encoder-decoder network is to predict the missing parts of incomplete point clouds. However,  we can only get predictions for missing proxies from the Transformer decoder. Therefore, we propose a multi-scale point cloud generation framework to recover missing point clouds at full resolution. To reduce redundant computations, we reuse the $M$ coordinates produced by the query generator as the local centers of the missing point cloud. Then, we utilize a reconstruction head like FoldingNet~\cite{FoldingNet} $f$ to recover detailed local shapes centered at the predicted proxies:
\begin{equation} \small
    {\mathcal{P}}_i = f(\mathcal{H}_i) + c_i, \quad i=1,2,...,M.
\end{equation}
where ${\mathcal{P}}_i$ is the set of neighboring points centered at $c_i$. Following previous work~\cite{PFNet}, we only predict the missing parts of the point cloud and \emph{concatenate} them with the input point cloud to obtain the complete objects. Both predicted proxies and recovered point clouds are supervised during the training process, and the detailed loss function will be introduced in the following section.

{\color{black}
\subsection{Adaptive Denoising Queries}  
\label{comp3}

The concatenation operation in $\mathcal{R}^3$ space is a simple solution for combining input points and predicted missing points, which treats these two parts of a point cloud as two separate units rather than a unified whole. The missing part is generated by a reconstruction head and the known part is obtained from the input (generated by the back-projecting method or captured by LiDAR sensors). However, this will bring some issues, like discontinuous and uneven appearance for the final predicted point cloud, which impairs the performance. One straightforward approach~\cite{CRN} to address this issue is to add some refinement modules after the concatenation and generate a new complete point cloud, which requires extra parameters and longer latency. Different from theirs, we propose AdaPoinTr (PoinTr+Adapative Denoising Queries), which takes the advantage of set-to-set translation formulation and point proxy representation, to address the issue by concatenating two sets of point proxies instead of concatenating two partial point clouds. The concatenated proxies are fed into a shared reconstruction model to produce the complete point clouds. In this way, the model reconstructs two parts in a unified manner and introduces no additional refinement parameters. Besides, we devise an advanced denoising task, which significantly improves the efficiency and robustness of our model. Two components of Adaptive Denoising Queries: 1) Adaptive queries generation mechanism and 2) Auxilliary denoising task are introduced in the following paragraphs.

\noindent \textbf{Adaptive Query Generation.} We modify the design of the original Query Generator and generate a dynamic query bank $\mathcal{B}$ conditioned on the encoder outputs $\mathcal{V}$ and the input point proxies $\mathcal{F}$, as shown in the Fig.~\ref{fig:query}(c). The dynamic query bank contains two types of queries: (a) $\mathcal{Q_I}$, abstracted from the input point proxies $\mathcal{F}$. (b) $\mathcal{Q_O}$, generated to serve as the initial state of the missing point proxies. Specifically, we summarize $\mathcal{V}$ and $\mathcal{F}$ with a linear projection to higher dimensions followed by the max-pooling, separately. Then the summarized features are projected into $M_I\times 3$ and $M_O\times 3$ dimension space and then reshaped as $M_I$ and $M_O$ coordinates $\mathcal{C^I}  = \{c_1^I, c_2^I, ..., c_{M_I}^I\} $ and $\mathcal{C^O} = \{c_1^O, c_2^O, ..., c_{M_O}^O\}$. After concatenating coordinates and the corresponding summarized features, we use an MLP to produce queries, which can be written as:
\begin{eqnarray} 
\nonumber
\mathcal{C^I} = \mathcal{P_I}(\mathcal{M}(\text{W}_I(\mathcal{F})), \quad
\mathcal{Q_I} = \text{MLP}([\mathcal{C^I}, \mathcal{M}(\text{W}_I(\mathcal{F})]), \\
\nonumber
\mathcal{C^O} = \mathcal{P_O}(\mathcal{M}(\text{W}_O(\mathcal{V})), \quad
\mathcal{Q_O} = \text{MLP}([\mathcal{C^O}, \mathcal{M}(\text{W}_O(\mathcal{V})]),
\end{eqnarray}%
where $\mathcal{M}$, $\mathcal{P_I}$, $\mathcal{P_O}$, $\text{W}_I$ and $\text{W}_O$ are max-pooling operation, coordinates projection for two sets and linear projection for two sets, respectively.
We further perform a query selection scheme to adaptively pick a portion of queries out. In this selection, we only constrain the total number of queries after selection without caring about the specific number of queries from $\mathcal{Q_I}$ and $\mathcal{Q_O}$, which makes our method more flexible in diverse situations. Technically, we design a light-weight scoring module $\mathcal{S}$ to rank the queries in $\mathcal{B}$ according to the predicted score and choose $M$ queries with higher scores, represented as $\mathcal{Q}$ with their coordinates $\mathcal{C}$. 

\paragrapha{2pt}{Auxiliary Denoising Task.} 
The Transformer decoder utilizes self-attention and cross-attention mechanisms to enhance the structural knowledge. It builds the relationship between queries and output features from the encoder to predict the missing point proxies. However, we observe an extremely unstable training curve during the early stage of model optimization (Shown in Fig.~\ref{fig:improvement}(a)) caused by the low-quality queries initialization. Although we propose a multi-scale point cloud generation scheme to directly supervise the coordinates $\mathcal{C}$ corresponding to queries before feeding them into the decoder, it can only alleviate the problem a little. We further design a denoising task by feeding some newly introduced noised queries along with the adaptive queries to the decoder, as shown in the Fig.~\ref{fig:query}(c). Specifically, we produce $k$ noised queries by adding a scaled random noise to $k$ groundtruth center points (obtained by FPS operation) in the input, $\mathcal{\hat{C}}^{gt} = \{\hat{c}_1^{gt}, \hat{c}_2^{gt}, ..., \hat{c}_k^{gt}\}$, where $\hat{c}_i^{gt} = n_i + c_i^{gt}$. The generation can be formulated as follows:
\begin{equation} \small
    \mathcal{\hat{Q}} = \text{MLP}([\mathcal{\hat{C}}, \mathcal{M}(\text{Linear}(\mathcal{V})]),
\end{equation}
which can be generated together with $\mathcal{Q_I}$ and $\mathcal{Q_O}$, since the MLP is shared for these sets of queries. This design benefits from two aspects. Firstly, it guarantees that there are always some high-quality queries fed into the decoder. Secondly, the denoising task improves the robustness of the decoder to initial queries. To avoid knowledge leakage from those noised queries $\mathcal{\hat{Q}}$ to $\mathcal{Q}$, we introduce an attention mask in self-attention layers.  Technically, we set the attention mask for noised queries $\mathcal{\hat{Q}}$ and $\mathcal{Q}$ to zero and keep the other attention unchanged, which can be written as follows:
\begin{equation}
\label{eq6}
m_{ij}=\left\{
\begin{aligned}
0 & , & \quad q_i \in \mathcal{\hat{Q}}, q_j \in \mathcal{Q}  \\
1 & , & otherwise.
\end{aligned}
\right.
\end{equation}
$\mathcal{\hat{Q}}$ is translated into predicted point proxies $\mathcal{\hat{H}}$ together with other normal queries $\mathcal{Q}$, which can be expressed as:
\begin{eqnarray} \small \nonumber
    \mathcal{[H, \hat{H}]} = \mathcal{M}_D( \mathcal{[Q, \hat{Q}]}, \mathcal{V}),
\end{eqnarray}
where $\mathcal{M}_D$ is Transformer decoder. Then, $\mathcal{\hat{H}}$ will be rebuilt as detailed local shapes centered at the predicted proxies:
\begin{equation} \small
    {\mathcal{\hat{P}}}_i = f(\hat{\mathcal{H}}_i) + \hat{c}^{gt}_i, \quad i=1,2,...,k.
\end{equation}
where ${\mathcal{\hat{P}}}_i$ is the set of neighboring points centered at $\hat{c}^{gt}_i$.

}

\subsection{Optimization}\label{loss}
The loss function for point cloud completion should provide a quantitative measurement for the quality of output. However, since the point clouds are unordered, many loss functions that directly measure the distance between two points (\ie $\ell_2$ distance) are unsuitable. Fan \etal~\cite{fan2017point} introduce two metrics that are invariant to the permutation of points, which are Chamfer Distance (CD) and Earth Mover's Distance (EMD). We adopt Chamfer Distance as our loss function for its $\mathcal{O}(N\log N)$ complexity. We use $\mathcal{C}$ to represent the ${n_{\mathcal{C}}}$ local centers and  $\mathcal{P}$ to represent ${n_{\mathcal{P}}}$ points of the complete point cloud. Give the ground-truth complete point cloud $\mathcal{G}$, the loss functions for these two predictions can be written as:
\begin{small}
\begin{eqnarray} \small
\label{equ:Loss_func_2_1}
\nonumber
J_0&=& \frac{1}{n_{\mathcal{C}}}\sum_{c\in \mathcal{C}} \min_{g\in \mathcal{G}} \|c-g\| + \frac{1}{n_{\mathcal{G}}}\sum_{g\in \mathcal{G}} \min_{c\in \mathcal{C}} \|g-c\|,    \\   
\nonumber
J_1&=& \frac{1}{n_{\mathcal{P}}}\sum_{p\in \mathcal{P}} \min_{g\in \mathcal{G}} \|p-g\| + \frac{1}{n_{\mathcal{G}}}\sum_{g\in \mathcal{G}} \min_{p\in \mathcal{P}} \|g-p\|.
\end{eqnarray}%
\end{small}%
 We directly use the high-resolution point cloud $\mathcal{G}$ to supervise the sparse point cloud $\mathcal{C}$ to encourage them to have similar distributions. {\color{black} As for the auxiliary denoising task, giving the noised queries $\mathcal{\hat{Q}}_i$ and the corresponding noised center $\hat{c}_i^{gt} = n_i + c_i^{gt}$, we expect our model can be robust to the random noise $n_i$ and rebuild detailed local shapes centered at $c_i^{gt}$. If we use $\mathcal{G}_{c_i}^{gt}$ to represent the ground-truth local shapes centered at $c_i^{gt}$ and use $\mathcal{\hat{P}}_i$ to represent the predicted local shapes by $\mathcal{\hat{Q}}_i$, the auxiliary loss function can be written as:
 \begin{equation} \small
 \nonumber
   J_{denoise} = \frac{1}{|\mathcal{\hat{P}}_i|}\sum_{c\in \mathcal{\hat{P}}_i} \min_{g \in \mathcal{G}_{c_i}^{gt}} \|c-g\| + \frac{1}{|\mathcal{G}_{c_i}^{gt}|}\sum_{g\in \mathcal{G}_{c_i}^{gt}} \min_{c\in \mathcal{\hat{P}}_i} \|g-c\|.
\end{equation}}
 Our final objective function for point cloud completion is the sum of these two objectives $J_{PC} = J_0 + J_1 + \lambda J_{denoise} $.

\begin{figure}[t]
    \centering
    \includegraphics[width = \linewidth]{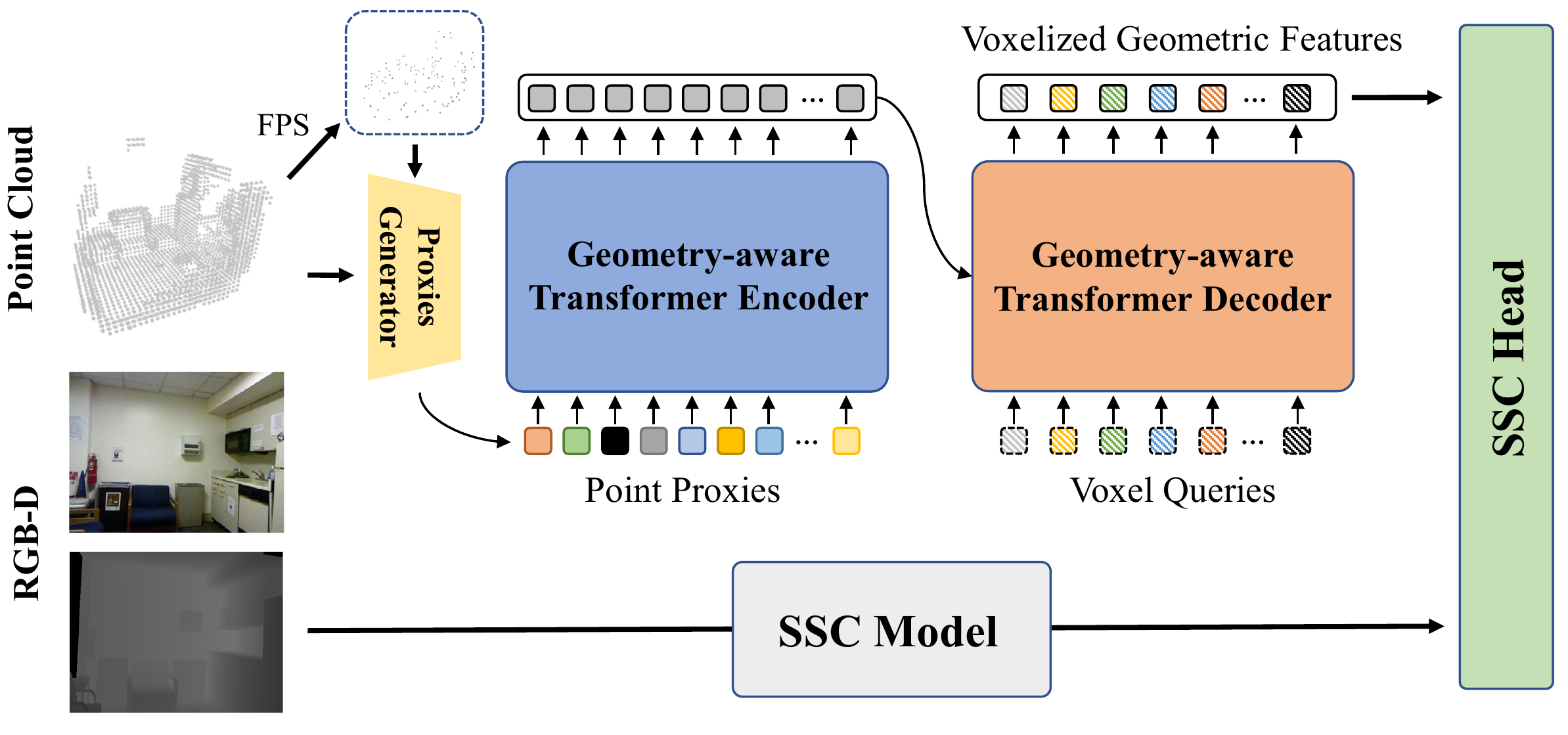}
    \caption{\small The proposed geometry-enhanced block {\color{blue}(top branch)} for semantic scene completion. The input point clouds for this block are obtained from the depth map. It acts as a plug-in block for any SSC models, capturing the detailed geometric information and performing global interactions for the whole scene.}
    \label{fig:GEBlock}
    \vspace{-10pt}
\end{figure}

{\color{black}
\section{PoinTr for Semantic Scene Completion}

{\color{black}
Taking a step towards scene-level completion is essential and meaningful for the rapidly growing 3D community. Compared with object-level point cloud completion, scene-level tasks bring more challenges, since interactions among the elements within a scene should be considered besides intra-object relations. {\color{black} 
We propose a new semantic scene completion framework and design a new \textit{Geometry-Enhanced block} based on PoinTr as a plug-and-play module for any semantic scene completion models. The overall framework is illustrated in Fig.~\ref{fig:GEBlock}}}. 

\vspace{-5pt}
\subsection{Problem Formulation}
Semantic Scene Completion (SSC) aims to simultaneously complete a 3D voxelized scene and predict the semantic labels of objects,  given a single-view depth map or RGB-D image. Each scene is voxelized into 3D volumes of 60$\times$36$\times$60 resolution, and the model is required to perform classification on each voxel in the scene:
{\small
$$
\hat{S} = \mathcal{M}_{SSC}(\mathcal{I}, \mathcal{D}),
$$}
\noindent where $\mathcal{I}$ and $\mathcal{D}$ represent the input images and depth maps, respectively. $\mathcal{M}_{SSC}$ is a general model for semantic scene completion. The output of the model, $\hat{S}$, represents the predicted probability over all classes for all voxels. In particular, $\hat{S} = \{\hat{s}_{i,j,k}\} \in \mathcal{R}^{\mathcal{N}+1}$, where $\mathcal{N}$ is the number of object categories and the first dimension of $\hat{s}_{i,j,k}$ represents the possibility of being an empty voxel. 

\subsection{Geometry-Enhanced Semantic Scene Completion}
There are two key challenges for semantic scene completion. The first issue is how to encode the inputs from 2D to 3D. Most recent work~\cite{song2017semantic, chen20203d, cai2021semantic, garbade2019two, li2019rgbd, liu2018see} utilizes depth maps to encode the geometric information of the seen surface and employs RGB images to compensate for semantic knowledge. Depth maps are encoded as Truncated Signed Distance Function (TSDF), where each voxel stores the distance $d$ to the closest surface with the sign indicating the voxel is free or occluded. This encoding method directly converts the 2D observation into 3D space and bridges the gaps between 2D inputs and 3D outputs. 
RGB images are processed with a separate branch for 2D feature maps and then projected into 3D voxelized space according to pixel-to-pixel correspondence with depth maps. Then, the encoded inputs are fed into a network consisting of 3D convolutions.

The second issue is how to compensate for the lack of geometric information in the occluded space, {\color{black} since an overall understanding of the whole scene including the occluded part is crucial to avoid the ambiguity caused by the incompleteness of some local areas
}. However, it still remains an open problem in this area. To explore the solutions to the above issue, we propose a new \textit{geometry-enhanced} semantic scene completion framework by inferring the absent geometric information from the incomplete scene and introducing it to the conventional SSC models, as shown in Fig.~\ref{fig:GEBlock}. We propose to convert the depth maps into 3D point clouds, and adopt a geometry-enhance block based on a modified PoinTr model to learn structural features, building the point-wise interactions for the whole scene. 
In our geometry-enhanced semantic scene completion framework, there are two main components: 1) conventional semantic scene model consisting of 3D convolutions; 2) geometry-enhanced block built with Transformers.

\vspace{-5pt}
\subsection{Point-to-Voxel Translation with Transformers}  
\label{sec:p2v}
Most previous work doesn't consider point clouds as an input stream since the final complete scenes are required to be in the voxel format. Based on point clouds, we seek a way to capture more precise geometric information and build point-wise interactions as a supplement to existing SSC models. We do not introduce any extra input data compared with other SSC models for the input point clouds are converted from depth maps. It is hard to directly adopt PoinTr for this task because the original PoinTr can not handle the point-to-voxel translation. Since the final scene is voxelized into a fixed size and the spatial location of each voxel is pre-defined, which runs counter to the \textit{Dynamic Queries} scheme in the original PoinTr, we modify the decoder of the original PoinTr to help it adapt to the SSC task. We pre-define $K$ \textit{Voxel Queries} as the inputs of the Transformer decoder instead of generating the queries in an online manner. Meanwhile, we use the same process to encode the incomplete scene as described in Sec.~\ref{sec:method}. Specifically, we represent the input point clouds of the scene with $S$ point proxies and use the Transformer encoder to build the long-range relationship among all the point proxies by the self-attention mechanism and then feed these outputs into the Transformer decoder. The Transformer decoder takes pre-defined \textit{Voxel Queries} and the outputs of the Transformer encoder as inputs and utilizes the attention mechanism to build the interactions between points and voxels to realize a point-to-voxel translation. After that, we feed the output of the decoder (i.e., \textit{Voxelized Geometric Features}) to other SSC models to provide the detailed geometric information for the entire pipeline.}

\subsection{Optimization}
\label{sec:optimize-scene}
{\color{black}
In semantic scene completion, the predicted scenes and the ground-truth scenes are voxelized before measuring the difference between two scenes. We follow the previous work~\cite{song2017semantic} to adopt the sum of voxel-wise cross-entropy loss as the loss function, which can be written as:
\begin{equation}
    \mathcal{J}_{SSC}(\hat{S}, S) = \sum_{i,j,k} \mathcal{L}_{\mathit{CE}(\hat{s}_{i,j,k}, s_{i,j,k})},
\end{equation}
\noindent where $\mathcal{L}_{\mathit{CE}}$ is cross-entropy loss, $\hat{s}_{i,j,k}$ and $s_{i,j,k}$ are the predicted probability over all classes and the ground-truth label of the voxel at coordinates (i, j, k), respectively. 
}

\begin{table*}[t]
\small 
\caption{\small Results on \textit{ShapeNet-55}. We report the detailed results for each method on 10 categories and the overall results on 55 categories for three difficulty degrees. $^\dag$ represents re-implemented results. We use CD-S, CD-M, and CD-H to represent the CD-$\ell_2$(multiplied by 1000) results under \emph{Simple}, \emph{Moderate}, and \emph{Hard} settings. We also provided F-Score@1\% averaged on three settings. } \vspace{-10pt}
\label{tab:ShapeNet-55}
\newcolumntype{g}{>{\columncolor{Gray}}c}
\centering
\setlength{\tabcolsep}{3pt}{
\adjustbox{width=\linewidth}{
\begin{tabular}{@{\hskip 5pt}>{\columncolor{white}[5pt][\tabcolsep]}l | c c c c c| c c c c c |c c c |g >{\columncolor{Gray}[\tabcolsep][5pt]}c@{\hskip 5pt}}
\toprule
&Table&Chair&Airplane&Car&Sofa&\tabincell{c}{Bird\\house}&Bag&Remote&\tabincell{c}{Key\\board}&Rocket& CD-S & CD-M & CD-H & CD-$\ell_2$-Avg & F-Score@1\%\\
\midrule
FoldingNet$^\dag$~\cite{FoldingNet}& 2.53 & 2.81 & 1.43  & 1.98&2.48 &4.71&2.79&1.44&1.24&1.48&2.67&2.66&4.05&3.12& 0.082\\
PCN$^\dag$~\cite{PCN}& 2.13 & 2.29 & 1.02 & 1.85&2.06&4.50&2.86&1.33&0.89&1.32&1.94&1.96&4.08&2.66 & 0.133\\
TopNet$^\dag$~\cite{TopNet}& 2.21 & 2.53 & 1.14  & 2.18&2.36&4.83&2.93&1.49&0.95&1.32&2.26&2.16&4.3&2.91 & 0.126\\
PFNet$^\dag$~\cite{PFNet}& 3.95  & 4.24  & 1.81  & 2.53&3.34&6.21&4.96&2.91 &1.29&2.36&3.83&3.87&7.97&5.22 & 0.339\\
GRNet$^\dag$~\cite{GRNet}& 1.63 & 1.88 & 1.02  & 1.64&1.72 &2.97&2.06&1.09&0.89&1.03&1.35&1.71&2.85&1.97 & 0.238\\
SnowflakeNet$^\dag$~\cite{xiang2021snowflakenet} & 0.98 & 1.12& 0.54 & 0.98 & 1.02 & 1.93 & 1.08 & 0.57 & 0.48 & 0.61 & 0.70 & 1.06 & 1.96&1.24 & 0.398\\
LAKeNet~\cite{tang2022lake} & -- & -- & -- & -- & -- & -- & -- & -- & -- & -- & -- & -- & -- & 0.89 & --\\
SeedFormer~\cite{zhou2022seedformer} & 0.72 & 0.81& 0.40 & 0.89 & 0.71 & -- & -- & -- & -- & -- & 0.50 & 0.77 & 1.49 & 0.92 & 0.472\\
\midrule
PoinTr~\cite{yu2021pointr}    &0.81  & 0.95  & 0.44  & 0.91 &0.79&1.86&0.93&0.53&0.38&0.57&0.58&0.88&1.79&1.09 & 0.464\\
\textbf{AdaPoinTr}   &\textbf{0.62}  &  \textbf{0.69}  & \textbf{0.33}  & \textbf{0.81} &\textbf{0.63}&\textbf{1.33}&\textbf{0.68}&\textbf{0.38}&\textbf{0.33}&\textbf{0.34}&\textbf{0.49}&\textbf{0.69}&\textbf{1.24}&\textbf{0.81} & \textbf{0.503}\\
\bottomrule
\end{tabular}}}
\end{table*}

\begin{table*}[t] \small 
\caption{\small Results on \textit{ShapeNet-34}. We report the results of 34 seen categories and 21 unseen categories in three difficulty degrees. $^\dag$ represents re-implemented results. We use CD-S, CD-M, and CD-H to represent the CD-$\ell_2$(multiplied by 1000) results under \emph{Simple}, \emph{Moderate}, and \emph{Hard} settings.  We also provided F-Score@1\% averaged on three settings.
} \vspace{-10pt}
\newcolumntype{g}{>{\columncolor{Gray}}c}
\label{tab:ShapeNet-34}
\centering
\setlength{\tabcolsep}{9pt}{
\adjustbox{width=\linewidth}{
\begin{tabular}{@{\hskip 5pt}>{\columncolor{white}[5pt][\tabcolsep]}lcccggcccg>{\columncolor{Gray}[\tabcolsep][5pt]}c@{\hskip 5pt}}
\toprule
\multirow{2}[1]{*}{} & \multicolumn{5}{c}{\textbf{34 seen categories}} & \multicolumn{5}{c}{\textbf{21 unseen categories}} \\
\cmidrule(lr){2-6}\cmidrule(lr){7-11} 
& CD-S& CD-M & CD-H & CD-$\ell_2$-Avg & F-Score@1\%  & CD-S & CD-M & CD-H & CD-$\ell_2$-Avg & F-Score@1\% \\
\midrule
FoldingNet$^\dag$~\cite{FoldingNet}& 1.86 & 1.81 & 3.38  & 2.35 & 0.139 &2.76&2.74&5.36&3.62&0.095\\
PCN$^\dag$~\cite{PCN}&1.87 & 1.81 & 2.97 & 2.22 & 0.154 &3.17&3.08&5.29&3.85&0.101\\
TopNet$^\dag$~\cite{TopNet}& 1.77 & 1.61 & 3.54  & 2.31 & 0.171  &2.62&2.43&5.44&3.50&0.121\\
PFNet$^\dag$~\cite{PFNet}& 3.16  & 3.19  & 7.71  & 4.68& 0.347  &5.29&5.87&13.33&8.16&0.322\\
GRNet$^\dag$~\cite{GRNet}& 1.26 & 1.39 & 2.57  & 1.74  & 0.251 &1.85&2.25&4.87&2.99&0.216\\
SnowflakeNet$^\dag$~\cite{xiang2021snowflakenet} & 0.60 & 0.86 & 1.50 & 0.99 &0.422 & 0.88 & 1.46 & 2.92 & 1.75 & 0.388 \\
SeedFormer~\cite{zhou2022seedformer} & \textbf{0.48} & 0.70 & 1.30 & 0.83 & 0.452 & \textbf{0.61} & 1.07 & 2.35 & 1.34 & 0.402\\
\midrule
PoinTr~\cite{yu2021pointr} &  0.76  & 1.05  & 1.88  &1.23 &0.421 &1.04&1.67&3.44&2.05&0.384\\
\textbf{AdaPoinTr}    &  \textbf{0.48}  & \textbf{0.63}  & \textbf{1.07}  & \textbf{0.73} & \textbf{0.469} & 0.61&\textbf{0.96}&\textbf{2.11}& \textbf{1.23} & \textbf{0.416}\\
\bottomrule
\end{tabular}}}
\end{table*}

\section{Experiments}

{\color{black} We conduct experiments on both object point cloud completion and semantic scene completion tasks.}  
In Section~\ref{sec:exp1}, we introduce the {\color{black}proposed} benchmarks for diverse point cloud completion and the evaluation metric. Then, we show the results of both our method and several baseline methods on our {\color{black} proposed} benchmarks. We also demonstrate the effectiveness of our model on the widely used PCN dataset, LiDAR-based KITTI benchmark, and MVP competition on ICCV 2021 Workshop. We perform the ablation studies to analyze each technical design in our AdaPoinTr. {\color{black} In Section~\ref{sec:exp2}, we detail the datasets and setups of scene-level completion and report the experimental results for semantic scene completion on NYUCAD~\cite{firman2016structured} and NYUV2~\cite{silberman2012indoor}.
}

\label{sec:experiment}
\subsection{Object Point Cloud Completion}
\label{sec:exp1}
\subsubsection{Benchmarks for Diverse Point Completion}
 We choose to generate the samples in our benchmarks based on the synthetic dataset, ShapeNet~\cite{ShapeNet}, because it contains the complete object models that cannot be obtained from real-world datasets like ScanNet~\cite{dai2017scannet} and S3DIS~\cite{armeni2017joint}. What makes our benchmarks distinct is that our benchmarks contain more object categories, more incomplete patterns, and more viewpoints. Besides, we pay more attention to the ability of networks to deal with the objects from novel categories that do not appear in the training set.

\paragrapha{3pt}{ShapeNet-55 Benchmark:} In this benchmark, we use all the objects in ShapeNet from 55 categories.  Most existing datasets for point cloud completion like PCN~\cite{PCN} only consider a relatively small number of categories (\eg, 8 categories in PCN). However, the incomplete point clouds from real-world scenarios are much more diverse. Therefore, we propose to evaluate the point cloud completion models on all 55 categories in ShapeNet to more comprehensively test the ability of models with a more diverse dataset.  We split the original ShapeNet using the 80-20 strategy: we randomly sample 80\% objects from each category to form the training set and use the rest for evaluation. As a result, we get 41,952 models for training and 10,518 models for testing. For each object, we randomly sample 8,192 points from the surface to obtain the point cloud.

\paragrapha{3pt}{ShapeNet-34 Benchmark:} In this benchmark, we want to explore another important issue in point cloud completion: the performance on novel categories. We believe it is necessary to build a benchmark for this task to better evaluate the generalization performance of models. We first split the origin ShapeNet into two parts: 21 unseen categories and 34 seen categories. In the seen categories, we randomly sample 100 objects from each category to construct a test set of the seen categories (3,400 objects in total) and leave the rest as the training set, resulting in 46,765 object models for training. We also construct another test set consisting of 2,305 objects from 21 novel categories. We evaluate the performance on both the seen and unseen categories to show the generalization ability of models.

\paragrapha{3pt}{Training and Evaluation:} In {\color{black} the \textit{ShapeNet-55} and \textit{ShapeNet-34}} benchmarks, the partial point clouds for training are generated online. We sample 2048 points from the object as the input and 8192 points as the ground truth. In order to mimic the real-world situation, we first randomly select a viewpoint and then remove the $n$ furthest points from the viewpoint to obtain a training partial point cloud. {\color{black} Our strategy for incomplete point clouds generation is more flexible and efficient {\color{black} for that we generate the incomplete point clouds in an online manner with little cost.}. 
} Besides, our strategy also ensures the diversity of our training samples in the aspect of viewpoints. During training, $n$ is randomly chosen from 2048 to 6144 (25\% to 75\% of the complete point cloud), resulting in different levels of incompleteness. We then down-sample the remaining point clouds to 2048 points as the input data for models.

During the evaluation, we fix 8 viewpoints and $n$ is set to 2048, 4096 or 6144 (25\%, 50\% or 75\% of the whole point cloud) for convenience. According to the value of $n$, we divide the test samples into three difficulty degrees, \textit{simple}, \textit{moderate}, and \textit{hard} in our experiments. In the following experiments, we will report the performance for each method in \textit{simple}, \textit{moderate}, and \textit{hard} to show the ability of each network to deal with tasks at difficulty levels. In addition, we use the average performance under three difficulty degrees to report the overall performance (\textit{Avg}). 

{\color{black}
\paragrapha{3pt}{Projected-ShapeNet-55/34 Benchmark:}
Considering to further bridge the gaps between incomplete point clouds in benchmarks and real-world scenarios, we propose \textit{Projected-ShapeNet-55} and \textit{Projected-ShapeNet-34}, which are another versions of original ShapeNet-55 and ShapeNet-34 sharing the basic setting except for incomplete point cloud generation. In original ShapeNet-55 and ShapeNet-34, the incomplete point clouds are generated by directly cropping complete point clouds, while the incomplete point clouds in \textit{Projected-ShapeNet-55} and \textit{Projected-ShapeNet-34} are generated by noised back-projecting depth images from one viewpoint, as shown in the Fig.~\ref{fig:dataset}. We randomly choose 16 viewpoints for each sample to render the depth image and obtain 16 different incomplete-complete point cloud pairs by performing noised back-projecting.
}

\begin{table*}[t]
\small 
\caption{\small Results on \textit{Projected-ShapeNet-55}. We report the detailed results for each method on 10 categories and the overall results on 55 categories under CD-$\ell_1$(multiplied by 1000). We also report F-Score@1\% metric. $^\dag$ represents re-implemented results. }
\label{tab:Projected_ShapeNet-55}
\newcolumntype{g}{>{\columncolor{Gray}}c}
\centering
\setlength{\tabcolsep}{8pt}{
\adjustbox{width=\linewidth}{
\begin{tabular}{@{\hskip 5pt}>{\columncolor{white}[5pt][\tabcolsep]}l | c c c c c| c c c c c |g>{\columncolor{Gray}[\tabcolsep][5pt]}c@{\hskip 5pt}}
\toprule
&Table&Chair&Airplane&Car&Sofa&\tabincell{c}{Bird\\house}&Bag&Remote&\tabincell{c}{Key\\board}&Rocket & CD-$\ell_1$ & F-Score@1\%\\
\midrule
PCN$^\dag$~\cite{PCN}& 14.79 & 15.33 & 9.07 & 12.85 & 17.12 & 20.38 & 18.64 & 14.62 & 13.69&10.98&16.64&0.403\\
TopNet$^\dag$~\cite{TopNet}& 14.40 & 16.29 & 9.85 & 13.61 & 16.93 & 22.00 & 18.69 & 13.52&11.05&10.45&16.35&0.337\\
GRNet$^\dag$~\cite{GRNet}& 12.01 & 12.57 & 8.30 & 12.13 & 14.36 & 16.52 & 14.67 & 12.18&9.71&8.58&12.81&0.491\\
SnowflakeNet$^\dag$~\cite{xiang2021snowflakenet} &  10.49 & 11.07 &6.35 & 11.20 & 12.59 & 15.24 & 12.86 & 10.07 & 8.12&7.49&11.34&0.594\\
\midrule
PoinTr~\cite{yu2021pointr} & 9.97 & 10.43 & 6.02 & 10.58 & 12.11 & 14.60 & 12.15& 9.55&7.61&6.86&10.68&0.615\\
AdaPoinTr & \textbf{8.81} & \textbf{9.12} & \textbf{5.18} & \textbf{9.77} & \textbf{10.89} & \textbf{13.27} & \textbf{10.93} & \textbf{8.81} & \textbf{6.79} & \textbf{5.58}&\textbf{9.58}&\textbf{0.701}\\
\bottomrule
\end{tabular}}}
\end{table*}

\begin{table*}[h!] \small 
\caption{\small Results on \textit{Projected-ShapeNet-34}. We report the results under CD-$\ell_1$(multiplied by 1000) of 34 seen categories and 21 unseen categories. We also report F-Score@1\% metric for each method. $^\dag$ represents re-implemented results.
} 
\newcolumntype{g}{>{\columncolor{Gray}}c}
\label{tab:Projected_ShapeNet-34}
\centering
\setlength{\tabcolsep}{7pt}{
\adjustbox{width=\linewidth}{
\begin{tabular}{@{\hskip 5pt}>{\columncolor{white}[5pt][\tabcolsep]}lccc|ggccc|g>{\columncolor{Gray}[\tabcolsep][5pt]}c@{\hskip 5pt}}
\toprule
\multirow{2}[1]{*}{} & \multicolumn{5}{c}{\textbf{34 seen categories}} & \multicolumn{5}{c}{\textbf{21 unseen categories}} \\
\cmidrule(lr){2-6}\cmidrule(lr){7-11} 
& Bin & Knife & Table & CD-$\ell_1$ & F-Score@1\%  & Microphone & Skateboard & Earphone & CD-$\ell_1$ & F-Score@1\% \\
\midrule

PCN$^\dag$~\cite{PCN}&  17.60 & 8.52 & 13.69 & 15.53 &0.432 & 18.05 & 17.27&24.82&21.44 &0.307\\
TopNet$^\dag$~\cite{TopNet}& 14.90 & 7.58 & 11.18 & 12.96 & 0.464 & 14.34&12.59&19.34&15.98 & 0.358\\
GRNet$^\dag$~\cite{GRNet} & 14.79 & 7.84 & 11.00&12.41&0.506&11.39&10.60&15.00&15.03&0.439\\
SnowflakeNet$^\dag$~\cite{xiang2021snowflakenet} &13.21 & 5.80 & 9.46&10.69&0.616&10.10&9.58&15.19&12.82&0.551\\
\midrule
 PoinTr~\cite{yu2021pointr}    &  12.36 & 5.64& 8.97& 10.21&0.634&9.34&8.98&14.23&12.43&0.555\\
 AdaPoinTr    &  \textbf{11.45} & \textbf{4.95}& \textbf{7.95}&\textbf{9.12}&\textbf{0.721}&\textbf{7.96}&\textbf{8.34}&\textbf{12.30}&\textbf{11.37}&\textbf{0.642}\\
\bottomrule
\end{tabular}}}
\end{table*}

\subsubsection{Evaluation Metric}
We follow the existing work~\cite{PCN,TopNet,PFNet,GRNet} to use the mean Chamfer Distance as the evaluation metric, which can measure the distance between the prediction point cloud and ground-truth in set-level. For each prediction, the Chamfer Distance between the prediction point set $\mathcal{P}$ and the ground-truth point set $\mathcal{G}$ is calculated by:
\begin{small}
$$
d_{CD}(\mathcal{P},\mathcal{G}) = \frac{1}{|\mathcal{P}|}\sum_{p\in \mathcal{P}} \min_{g\in \mathcal{G}} \|p-g\| + \frac{1}{|\mathcal{G}|}\sum_{g\in \mathcal{G}} \min_{p\in \mathcal{P}} \|g-p\|.
$$
\end{small}

\noindent Following the previous methods, we use two versions of Chamfer distance as the evaluation metric to compare the performance with existing work. CD-$\ell_1$ uses L1-norm to calculate the distance between two points, while CD-$\ell_2$ uses L2-norm instead. We also follow~\cite{FScore} to adopt F-Score as another evaluation metric. We define the precision and recall for a point cloud completion result at the threshold $d$ as:
\vspace{-5pt}
\begin{small}
\begin{eqnarray}
\nonumber
P(d) =  \frac{1}{|\mathcal{P}|}\sum_{p\in \mathcal{P}}  [\min_{g\in \mathcal{G}}  ||p-g|| < d], \\
\nonumber
R(d) = \frac{1}{|\mathcal{G}|}\sum_{g\in \mathcal{G}} \lbrack \min_{p\in \mathcal{P}} ||g-p|| < d].
\end{eqnarray}
\end{small}

\noindent Then we can calculate the F-Score@d by: 
\begin{small}
$$
\text{F-Score}(d) = \frac{2P(d)R(d)}{P(d)+R(d)}, 
$$
\end{small}

\noindent where P(d) and R(d) denote the precision and recall. In our experiments, we set the threshold $d$ to 1\%.

\subsubsection{Results on ShapeNet-55}
We report the performance of our model PoinTr and AdaPoinTr in Table~\ref{tab:ShapeNet-55} and compare them with the existing methods. We implement FoldingNet~\cite{FoldingNet}, PCN~\cite{PCN}, TopNet~\cite{TopNet}, PFNet~\cite{PFNet}, GRNet~\cite{GRNet} and SnowflakeNet~\cite{xiang2021snowflakenet} on our benchmark according to their open-source code, using the best hyper-parameters in their papers for fair comparisons.
As shown in the table, our method can better cope with different situations with diverse viewpoints, diverse object categories, diverse incomplete patterns, and diverse incompleteness levels. We report the average chamfer distance on three settings for ten categories in the first ten columns of the table. Specifically, \textit{Table}, \textit{chair}, \textit{Airplane},\textit{Car} and \textit{Sofa} contain more than 2500 samples in the training set while \textit{Birdhouse}, \textit{Bag}, \textit{Remote}, \text{Keyboard} and \textit{Rocket} contain less than 80 samples. And we also provide detailed results for all 55 categories in our supplementary material. As shown in the last two columns of the table, our AdaPoinTr performs better than other existing work and establish a new state-of-the-art, which achieves {\color {black} 0.81 CD-$\ell_2$ and 0.503} F-Score on ShapeNet-55. These results clearly demonstrate the effectiveness of our AdaPoinTr even under such a diverse situation.
{\color{black} Besides, AdaPoinTr achieves 0.09, 0.19 and 0.55 improvement in CD-$\ell_2$ (multiplied by 1000) under three settings (simple, moderate and hard) compared with PoinTr.}

\subsubsection{Results on ShapeNet-34}
\begin{figure}[t]
  \centering
  \includegraphics[width = \linewidth]{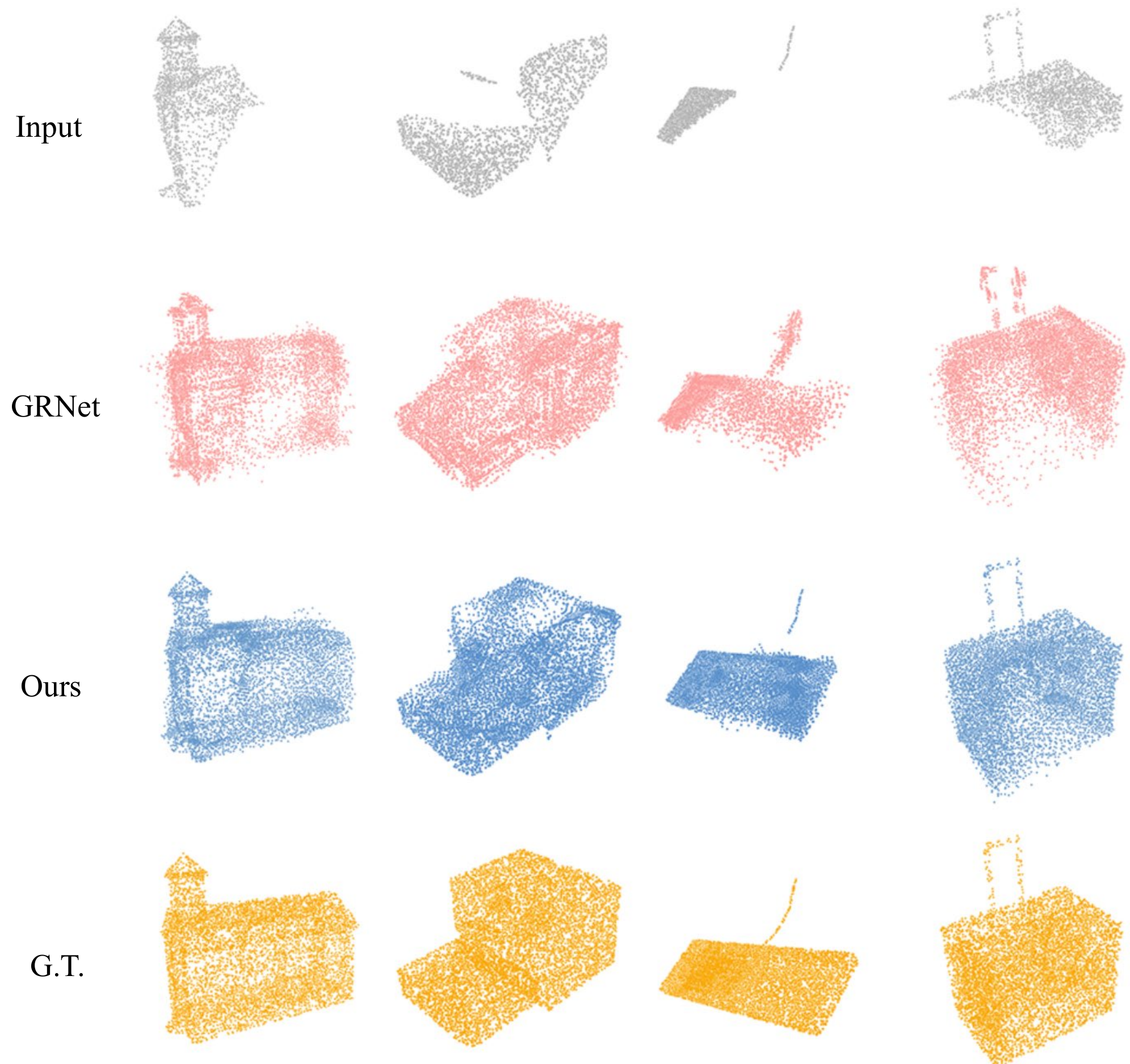}
  \caption{\small Results on some objects from novel categories in ShapeNet-34. We show the input point cloud and the ground truth as well as the predictions of GRNet and our PoinTr. }
  \label{fig:novel}
\end{figure}

On ShapeNet-34, we also conduct experiments for our method and existing methods. The results are shown in Table~\ref{tab:ShapeNet-34}. For the 34 seen categories, we can see our method outperforms all the other methods. For the 21 unseen categories, we use the networks that are trained on the 34 seen categories to evaluate the performance on the novel objects from the other 21 categories that do not appear in the training phase. We see our method also achieves the best performance in this more challenging setting. Comparing with the results of seen categories, {\color{black} our AdaPoinTr obtains a more dominant performance over different difficulty degrees, which demonstrates the superior transferability of our AdaPoinTr.} We visualize the results in Fig.~\ref{fig:novel} to show the effectiveness of our method on the unseen categories.

\begin{table*}[t]
\caption{Results on LiDAR scans from KITTI dataset under the  Fidelity and MMD metrics. We follow the previous work to finetune the model on PCNCars.} 
\vspace{-5pt}
\label{tab:KITTI}
\centering
\newcolumntype{g}{>{\columncolor{Gray}}c}
\begin{adjustbox}{width=\textwidth}
\begin{tabular}[\linewidth]{l | c c c c c c c c c c | g g}
\toprule
CD-$\ell_2$ ($\times$ 1000) & AtlasNet~\cite{AtlasNet} & PCN~\cite{PCN} & FoldingNet~\cite{FoldingNet} & TopNet~\cite{TopNet} & MSN~\cite{MSN} & NSFA~\cite{zhang2020detail} & PFNet~\cite{PFNet} &   CRN~\cite{CRN} & GRNet~\cite{GRNet} & SeedFormer & PoinTr & AdaPoinTr\\
\midrule
Fidelity $\downarrow$ & 1.759 & 2.235 & 7.467 &  5.354 &  0.434 & 1.281 & 1.137  & 1.023 & 0.816 & 0.151& \textbf{0.000} & 0.237 \\
MMD $\downarrow$ & 2.108 & 1.366 & 0.537 & 0.636 & 2.259 &0.891 & 0.792& 0.872& 0.568 & 0.516  & 0.526 & \textbf{0.392} \\
\bottomrule
\end{tabular}
\end{adjustbox}
\end{table*}

\begin{table*}[t]
\caption{\small Results on the PCN dataset. We use the CD-$\ell_1$(multiplied by 1000) and F-Score@1\% to compare with other methods.} 
\label{tab:PCN_benchmark}
\centering
\newcolumntype{g}{>{\columncolor{Gray}}c}
\begin{adjustbox}{width=\textwidth}
\begin{tabu}[0.95\linewidth]{l |*{8}{X[r]}|rr} 
\toprule
&  Air & Cab & Car & Cha & Lam & Sof & Tab & Wat  & Avg CD-$\ell_1$ & F-Score@1\% \\
\midrule
FoldingNet~\cite{FoldingNet} & 9.49 & 15.80 & 12.61 & 15.55 & 16.41 & 15.97 & 13.65 & 14.99 & 14.31&0.322\\
AtlasNet~\cite{AtlasNet}& 6.37& 11.94 &10.10& 12.06& 12.37 &12.99& 10.33 &10.61& 10.85&0.616 \\
PCN~\cite{PCN} & 5.50 & 22.70 & 10.63 & 8.70 & 11.00 & 11.34 & 11.68 & 8.59 & 9.64&0.695\\
TopNet~\cite{TopNet} & 7.61&  13.31&  10.90&  13.82&  14.44&  14.78&  11.22&  11.12& 12.15&0.503\\
MSN~\cite{MSN} &5.60& 11.90& 10.30 &10.20&10.70& 11.60&9.60&9.90 &10.00&0.705 \\
GRNet~\cite{GRNet} &6.45& 10.37 &9.45 &9.41& 7.96 &10.51 &8.44 &8.04 & 8.83 & 0.708\\
PMP-Net~\cite{PMP}& 5.65 &11.24& 9.64& 9.51 &6.95 &10.83& 8.72& 7.25&8.73&--\\
CRN~\cite{CRN} & 4.79&9.97& 8.31& 9.49& 8.94& 10.69& 7.81& 8.05&8.51&--\\
NSFA~\cite{zhang2020detail} & 4.76& 10.18& 8.63& 8.53& 7.03& 10.53& 7.35& 7.48 & 8.06& --\\
SnowFlake~\cite{xiang2021snowflakenet}& 4.29&9.16& 8.08& 7.89& 6.07& 9.23& 
6.55& 6.40&7.21& --\\
LAKeNet~\cite{tang2022lake}& 4.17 & 9.78& 8.56& 7.45& 5.88& 9.39& 
6.43& 5.98 &7.23& --\\
SeedFormer~\cite{zhou2022seedformer}& 3.85&9.05& 8.06& 7.06& \textbf{5.21}& 8.85& 
6.05& 5.85&6.74& --\\
\midrule
PoinTr  &4.75&10.47&8.68&9.39&7.75&10.93&7.78&7.29 &8.38& 0.745 \\
AdaPoinTr   & \textbf{3.68}&\textbf{8.82}&\textbf{7.47}&\textbf{6.85}&5.47&\textbf{8.35}&\textbf{5.80}&\textbf{5.76}&\textbf{6.53} &\textbf{0.845}\\

\bottomrule
\end{tabu}
\end{adjustbox}
\end{table*}

{\color{black}
\subsubsection{Results on Projected-ShapeNet-55}
We conduct experiments on Projected-ShapeNet-55, where the input point cloud is generated by the noised back-projecting method. In Projected-ShapeNet-55, there exists a noise in the input point cloud, making the dataset more challenging and realistic. We report the results of our models and other existing methods. We report the class-wise CD and overall CD for all methods. Limited by the page size, we only pick 10 categories of all to show the detailed results. As shown in Table~\ref{tab:Projected_ShapeNet-55}, our AdaPoinTr obtains the best performance on ten categories and overall CD, achieving 1.1 improvements on CD-$\ell_1$(multiplied by 1000) and 0.086 on F-Score@1\% compared with PoinTr. 

\subsubsection{Results on Projected-ShapeNet-34}
We also explore the performance of our AdaPoinTr and other existing methods on Project-ShapeNet-34, as shown in Table~\ref{tab:Projected_ShapeNet-34}. On Project-ShapeNet-34, the model is trained on the training split of 34 seen categories, then tested on the test split of 34 seen categories and 21 unseen categories. AdaPoinTr outperforms SnowflakeNet~\cite{xiang2021snowflakenet} on both 34 seen categories and 21 unseen 21 categories.
}

 \begin{figure*}[t]
  \centering
  \includegraphics[width = \linewidth]{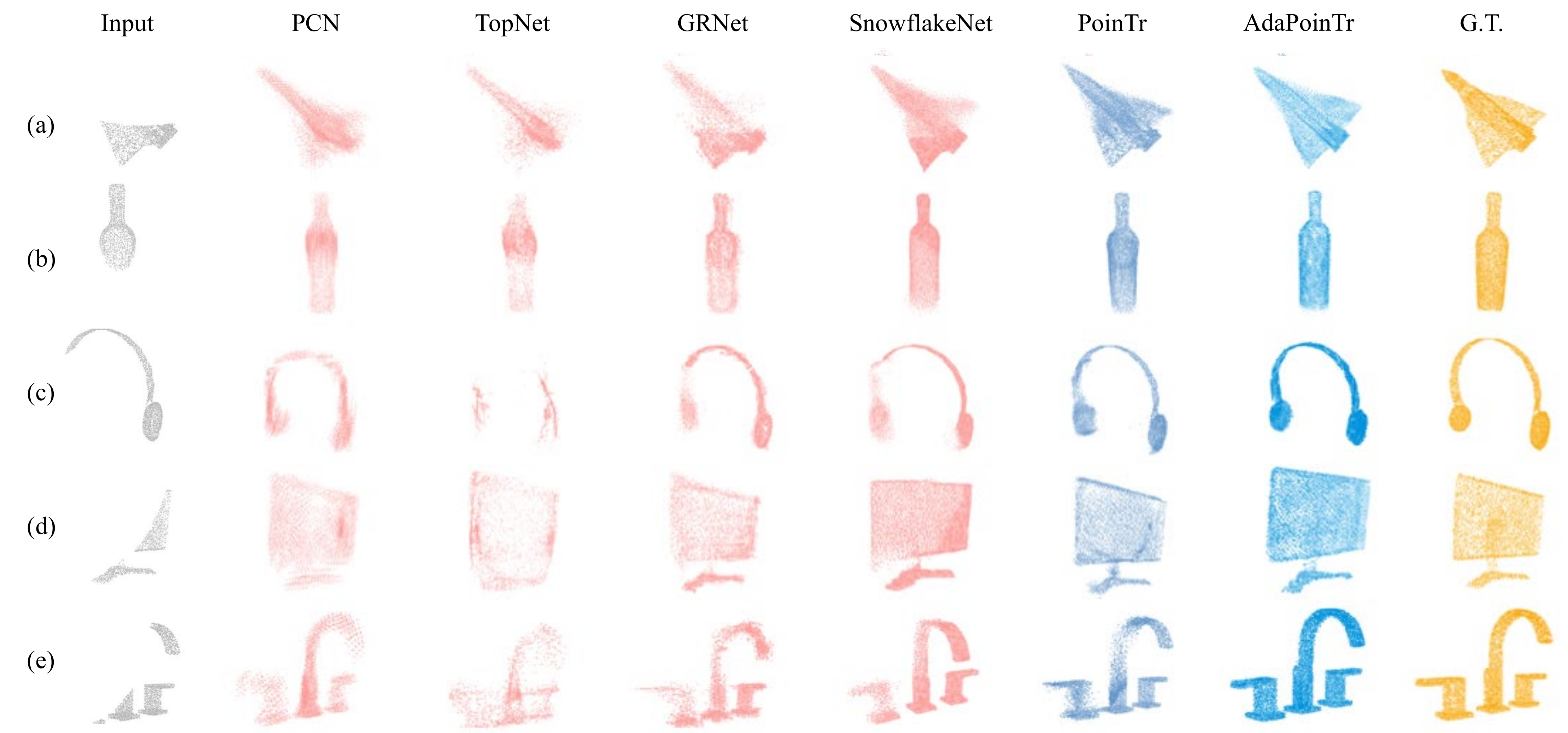}
  \caption{ Qualitative results on ShapeNet-55. All methods above take the point clouds in the first column as inputs and generate complete point clouds. Our methods can complete the point clouds with higher fidelity, which clearly shows the effectiveness of our method.} \vspace{-10pt}
  \label{fig:case}
\end{figure*}

 \begin{figure}[t]
  \centering
  \includegraphics[width = \linewidth]{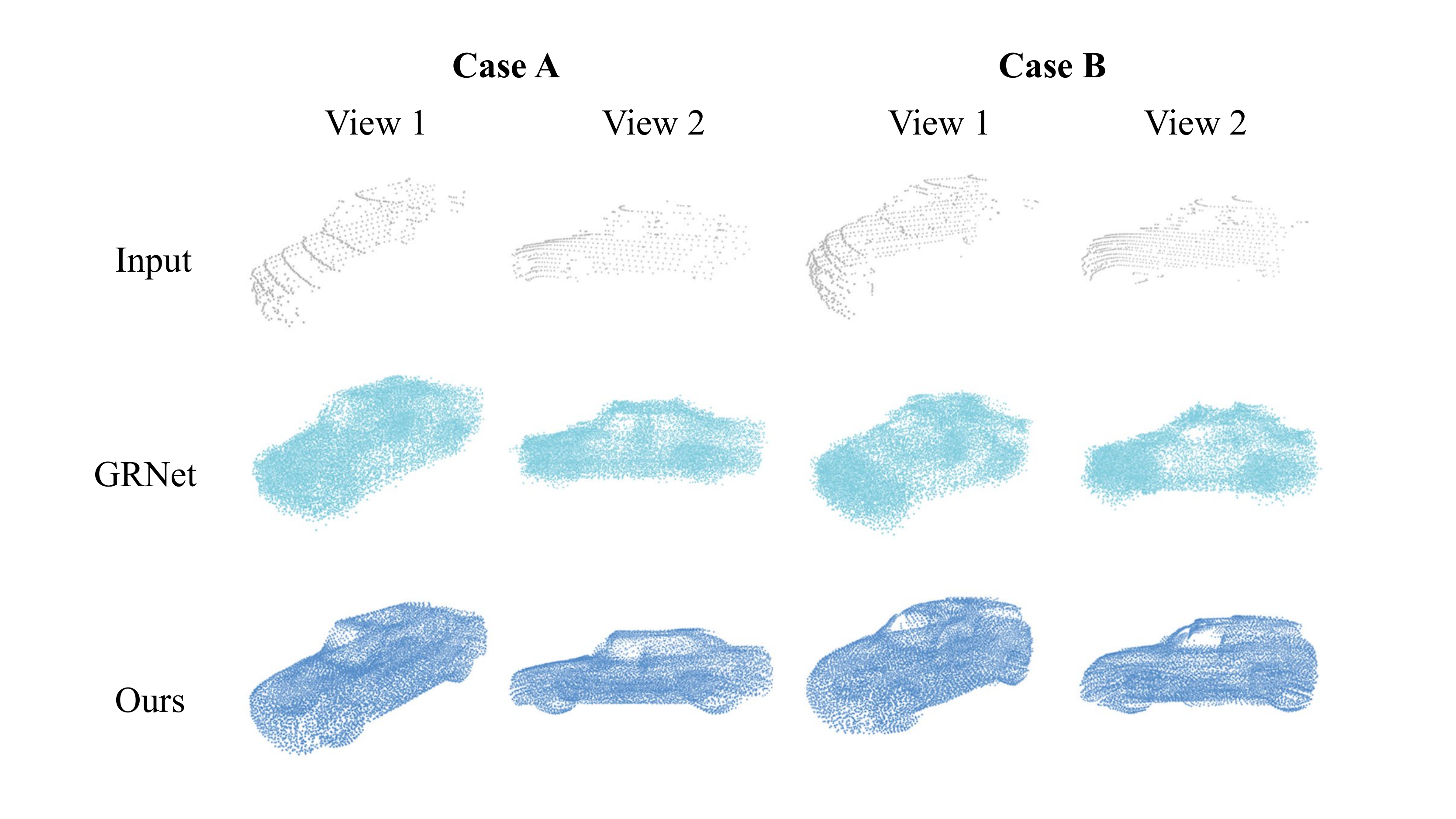}
  \caption{\small Qualitative results on the KITTI. In order to better show the shape of the car, we provide two views of the same point cloud in each case. Our method can recover a car with more accurate boundaries and details (\eg tires of cars).}
  \label{fig:KITTI}
\end{figure}

\begin{table}[t]
\small
\caption{\small Results on the MVP validation set. Inputs and outputs contain 2048 points. We report CD-$\ell_2$(multiplied by 10000) and F-Score@1\% for each method.} 
\centering
\label{tab:mvp}
\setlength{\tabcolsep}{15pt}{
\adjustbox{width=\linewidth}{
\begin{tabular}{@{\hskip 5pt}>{\columncolor{white}[5pt][\tabcolsep]}lcc>{\columncolor{white}[\tabcolsep][5pt]}c@{\hskip 5pt}}
\toprule
Model  &  \#Points & CD-$\ell_2$ & F-Score@1\%\\
\midrule
PCN~\cite{PCN}   & 2048 & 9.77  & 0.321 \\
TopNet~\cite{TopNet} & 2048 & 10.11 & 0.308 \\
ECG~\cite{ecg} & 2048  & 7.25  & 0.434 \\
CRN~\cite{CRN}  & 2048 & 6.64  & 0.476 \\
VRCNet~\cite{vrc} & 2048 & 5.96  & 0.499 \\
\midrule
\rowcolor{Gray}   PoinTr & 2048 & 6.15  & 0.456 \\
\rowcolor{Gray}  AdaPoinTr & 2048  & \textbf{4.71} & \textbf{0.545} \\
\bottomrule
\end{tabular}}}
\end{table}

\subsubsection{Results on the Existing Benchmarks}
Apart from the experiments on the newly proposed challenging benchmarks, we also conduct the experiments on the existing benchmarks including the PCN dataset~\cite{PCN}, LiDAR-based KITTI benchmark~\cite{KITTI} and MVP Challenges~\cite{vrc}.

\paragrapha{5pt}{Results on the PCN Dataset. } The PCN dataset~\cite{PCN} is one of the most widely used benchmark datasets for the point cloud completion task. To verify the effectiveness of our method on existing benchmarks and compare it with more state-of-the-art methods, we conducted experiments on this dataset following the standard protocol and evaluation metric used in previous work~\cite{PCN,MSN,GRNet,PMP,CRN,xiang2021snowflakenet}. The results are shown in Table~\ref{tab:PCN_benchmark}. We see our method largely improves the previous methods and establishes the new state-of-the-art on this dataset.

\paragrapha{5pt}{Results on the KITTI Benchmark. } To show the performance of our method in real-world scenarios, we follow~\cite{GRNet} to finetune our trained model on ShapeNetCars~\cite{PCN} and evaluate the performance of our model on KITTI dataset, which contains the incomplete point clouds of cars in the real-world scenes from LiDAR scans. We report the Fidelity and MMD metrics in Table~\ref{tab:KITTI} and show some reconstruction results in Fig.~\ref{fig:KITTI}. Our method achieves better qualitative and quantitative performance. See supplementary for the description of Fidelity and MMD.

{\color{black}
\paragrapha{5pt}{Results on the MVP Benchmark.}
The MVP benchmark evaluates the performance of generating complete 3D point clouds based on single-view partial point clouds. We conduct the experiments on its validation set and submit our model to MVP Challenges hosted in the ICCV 2021 Workshop for the online evaluation on the test set. We show the results on validation set in Table~\ref{tab:mvp}. Our method achieves the best performance among all the previous work, including the strong SOTA method VRCNet~\cite{vrc}. Besides, we ranked first on the online leaderboard of the MVP benchmark and won the first prize in the MVP Challenge~\cite{pan2021multi}. 
}

\begin{table}[t]
\small
\caption{\small Ablation study on the PCN dataset. We investigate different designs including query generator (Query), DGCNN feature extractor (DGCNN), Geometry-aware Blocks (Geometry), Adaptive Query Generation (Ada. Query), Query Selection (Selection) and Denoising task (Denoising). We report CD-$\ell_1$(multiplied by 1000) and F-Score@1\%.} 
\label{tab:Ablation Study}
\centering
\newcolumntype{g}{>{\columncolor{Gray}}c}

\setlength{\tabcolsep}{3pt}{
\adjustbox{width=\linewidth}{
\begin{tabular}{@{\hskip 5pt}>{\columncolor{white}[5pt][\tabcolsep]}c | c c c |g>{\columncolor{Gray}[\tabcolsep][5pt]}c@{\hskip 5pt}}

\toprule
Model  & Query & DGCNN & Geometry & CD-$\ell_1$ & F-Score@1\%\\
\midrule
A & & & &  9.43 & 0.678 \\
B &\checkmark & & & 9.09 & 0.713 \\
C    &\checkmark &\checkmark & &  8.69 & 0.736 \\
D     &\checkmark &\checkmark & all& 8.44 & 0.741 \\
PoinTr    &\checkmark &\checkmark & $1^{\rm st}$ & 8.38 & 0.745 \\
\midrule
Model  & Ada. Query & Selection & Denoising & CD-$\ell_1$ & F-Score@1\%\\
\midrule
PoinTr & & & &  8.38 & 0.745 \\
F &\checkmark & & & 7.06 & 0.797 \\
G    &\checkmark &\checkmark & &  6.92 & 0.810 \\
AdaPoinTr    &\checkmark &\checkmark & \checkmark & 6.53 & 0.844 \\
\bottomrule
\end{tabular}}}
\end{table}

\begin{table}[t]
\small
\caption{Complexity analysis. We report theoretical computation cost (FLOPs) and throughput (T.put) of our method and existing methods. We also provide Chamfer distances metric of all categories in ShapeNet-55 and projected-ShapeNet-55 (CD$_{\rm 55}$ and CD$_{\rm p55}$), unseen categories in ShapeNet34 and projected-ShapeNet-34 (CD$_{\rm 34}$ and CD$_{\rm p34}$) and PCN benchmark as references.} 
\label{tab:FLOPs}
\centering
\setlength{\tabcolsep}{0.8mm}{
\begin{adjustbox}{width=0.5\textwidth}
\begin{tabular}[\linewidth]{l | r r | c c c c | c}
\toprule
Models & FLOPs & T.put & CD$_{\rm 55}$ &  CD$_{\rm 34}$ & CD$_{\rm p55}$ &  CD$_{\rm p34}$ & CD$_{\rm PCN}$\\
\midrule
FoldingNet~\cite{FoldingNet} &27.58 G & 158 pc/s &3.12 & 3.62 & -- & -- & 14.31 \\
PCN~\cite{PCN}      & 15.25 G & 292  pc/s & 2.66 & 3.85 & 16.64 & 21.44 & 9.64\\
TopNet~\cite{TopNet}   &6.72 G & 248 pc/s&2.91 & 3.50 & 16.35 & 15.98 & 12.15\\
GRNet~\cite{GRNet}    & 40.44 G & 65 pc/s&1.97 & 2.99 & 12.81 & 15.03 & 8.83\\
SnowflakeNet~\cite{xiang2021snowflakenet}  & 17.16 G & 72 pc/s & 1.24 & 1.75 & 11.34 & 12.82 & 7.21\\
LAKeNet~\cite{tang2022lake}  & 24.48 G & 3 pc/s & 0.89 & -- & -- & -- & 7.23\\
SeedFormer~\cite{zhou2022seedformer} & 13.97 G & 21 pc/s & 0.92 & 1.34 & -- & -- & 6.74\\
\midrule
PoinTr   & 10.41 G & 62 pc/s& 1.09 & 2.05 & 10.68 & 12.43 & 8.38\\
AdaPoinTr  & 12.43 G & 61 pc/s & \textbf{0.81} & \textbf{1.23} & \textbf{9.58} & \textbf{11.37} & \textbf{6.53}\\

\bottomrule
\end{tabular}
\end{adjustbox}
}
\end{table}

\begin{table*}[t] \small 
\caption{\small Results on NYUV2 dataset. We compare our method with other end-to-end and iterative methods. We evaluate the models using \textit{Precision}, \textit{Recall} and \textit{IoU} for Scene Completion, detailed \textit{IoU} and \textit{mIoU} for Semantic Scene Completion. PoinTr only takes depth maps as the input, while other methods take additional RGB images as another input.
} 
\newcolumntype{g}{>{\columncolor{Gray}}c}
\label{tab:NYUV2}
\centering
\setlength{\tabcolsep}{7pt}{
\begin{adjustbox}{width=\linewidth} \small
\begin{tabular}{l c ccg cccccccccccg}
\toprule
\multirow{2}[0]{*}{} & \multirow{2}[0]{*}{\textbf{Type}} & \multicolumn{3}{c}{\textbf{Scene Completion}} & \multicolumn{12}{c}{\textbf{Semantic Scene Completion}} \\
\cmidrule(lr){3-5}\cmidrule(lr){6-17} 
& & Prec. & Recall & IoU & ceil & floor & wall & win. & chair & bed & sofa & table & tvs & furn & objs & mIoU \\
\midrule
SISNet~\cite{cai2021semantic} & Iterative & 90.7 & 84.6 & 77.8 & 53.9&  93.2 & 51.3 & 38.0 & 38.7 & 65.0 & 56.3 & 37.8 & 25.9 & 51.3 & 36.0&  49.8\\
\midrule
PoinTr~\cite{yu2021pointr} & Depth Only & 71.2 & 91.3 & 62.9 & -- & -- & -- & -- & -- & -- & -- & -- & -- & -- & -- & --\\
\midrule
SSCNet~\cite{song2017semantic}& End-to-end & 57.0& 94.5& 55.1& 15.1 &94.7& 24.4& 0.0& 12.6& 32.1& 35.0& 13.0 &7.8& 27.1 &10.1 &24.7\\
AICNet~\cite{li2020anisotropic}& End-to-end & 62.4& 91.8& 59.2& 23.2& 90.8& 32.3& 14.8& 18.2& 51.1& 44.8& 15.2& 22.4 &38.3& 15.7& 33.3\\
TS3D~\cite{garbade2019two} & End-to-end & -- & -- & 60.0& 9.7& 93.4& 25.5& 21.0& 17.4& 55.9& 49.2& 17.0& 27.5& 39.4& 19.3& 34.1\\
CCPNet~\cite{zhang2019cascaded}& End-to-end & 74.2 &90.8& 63.5& 23.5& 96.3& 35.7& 20.2& 25.8& 61.4& 56.1& 18.1 &28.1 &37.8& 20.1 &38.5\\
SISNet-Base~\cite{cai2021semantic}& End-to-end &87.6 &78.9& 71.0 &46.9& 93.3 &41.3 &26.7 &30.8& 58.4 &49.5& 27.2& 22.1& 42.2 &28.7& 42.5\\
\midrule
DDRNet~\cite{li2019rgbd}& End-to-end & 71.5& 80.8& 61.0& 21.1 &92.2& 33.5& 6.8& 14.8 &48.3& 42.3& 13.2& 13.9& 35.3 &13.2& 30.4\\
Sketch~\cite{chen20203d}& End-to-end &85.0 &81.6& 71.3& 43.1& 93.6& 40.5& 24.3& 30.0 &57.1& 49.3& 29.2 &14.3& 42.5 &28.6 &41.1\\
\midrule
DDRNet-GE& End-to-end & 76.2 & 81.9 & 65.2 & 24.6 & 92.4& 36.4& 17.1& 18.7& 49.1& 41.9& 15.9& 27.6& 37.1& 15.6 & 34.2\\
Sketch-GE& End-to-end & 90.1 & 82.9 & \textbf{74.6} & 45.6 & 93.7& 41.5& 29.5& 35.4& 56.3& 48.1& 26.9& 32.8& 45.1& 30.3 & \textbf{44.1}\\

\bottomrule
\end{tabular}
\end{adjustbox}}
\end{table*}

\begin{table*}[t] \small 
\caption{\small Results on NYUCAD dataset. We compare our method with other end-to-end and iterative methods. We evaluate the models using \textit{Precision}, \textit{Recall} and \textit{IoU} for Scene Completion, detailed \textit{IoU} and \textit{mIoU} for Semantic Scene Completion. PoinTr only takes depth maps as the input, while other methods take additional RGB images as another input.
} 
\newcolumntype{g}{>{\columncolor{Gray}}c}
\label{tab:NYUCAD}
\centering
\setlength{\tabcolsep}{7pt}{
\begin{adjustbox}{width=\linewidth} \normalsize
\begin{tabular}{l c ccg cccccccccccg}
\toprule
\multirow{2}[0]{*}{} & \multirow{2}[0]{*}{\textbf{Type}} & \multicolumn{3}{c}{\textbf{Scene Completion}} & \multicolumn{12}{c}{\textbf{Semantic Scene Completion}} \\
\cmidrule(lr){3-5}\cmidrule(lr){6-17} 
& & Prec. & Recall & IoU & ceil & floor & wall & win. & chair & bed & sofa & table & tvs & furn & objs & mIoU \\
\midrule
SISNet~\cite{cai2021semantic} & Iterative & 94.2& 91.3& 86.5& 65.6& 94.4& 67.1& 45.2& 57.2& 75.5& 66.4& 50.9& 31.1& 62.5& 42.9& 59.9\\
\midrule
PoinTr~\cite{yu2021pointr} & Depth Only & 89.8 & 91.7 & 80.9 & -- & -- & -- & -- & -- & -- & -- & -- & -- & -- & -- & --\\
\midrule
SSCNet~\cite{song2017semantic}& End-to-end & 75.4 & 96.3&  73.2&  32.5&  92.6&  40.2&  8.9&  33.9&  57.0&  59.5&  28.3&  8.1&  44.8&  25.1&  40.0\\
AICNet~\cite{li2020anisotropic}& End-to-end & 88.2& 90.3& 80.5& 53.0& 91.2& 57.2& 20.2& 44.6& 58.4& 56.2& 36.2& 9.7& 47.1& 30.4& 45.8\\
TS3D~\cite{garbade2019two} & End-to-end & -- & -- & 76.1& 25.9& 93.8& 48.9& 33.4& 31.2& 66.1& 56.4& 31.6& 38.5& 51.4& 30.8& 46.2\\
CCPNet~\cite{zhang2019cascaded}& End-to-end & 91.3& 92.6& 82.4& 56.2& 94.6& 58.7& 35.1& 44.8& 68.6& 65.3& 37.6& 35.5& 53.1& 35.2& 53.2\\
SISNet-Base~\cite{cai2021semantic}& End-to-end & --&--&82.8 & --& --& --& --& --& --& --& --& --& --& --& 53.6\\
\midrule
DDRNet~\cite{li2019rgbd}& End-to-end & 88.7 & 88.5& 79.4& 54.1& 91.5& 56.4& 14.9& 37.0& 55.7& 51.0& 28.8& 9.2& 44.1& 27.8& 42.8\\
Sketch~\cite{chen20203d}& End-to-end &90.6& 92.2& 84.2& 59.7& 94.3& 64.3& 32.6& 51.7& 72.0& 68.7& 45.9& 19.0& 60.5& 38.5& 55.2\\
\midrule
DDRNet-GE& End-to-end & 89.4 & 90.2 & 81.2 & 56.7& 91.9 & 57.7& 25.1& 36.5& 54.1& 49.3& 29.1& 21.8& 43.2& 26.9 & 44.7\\
Sketch-GE& End-to-end & 94.4 & 90.64 & \textbf{86.1} & 65.1&94.4& 64.9 & 40.9 & 50.3& 69.4& 58.5& 38.5& 40.8&57.7& 37.2 & \textbf{56.1}\\
\bottomrule
\end{tabular}
\end{adjustbox}}
\end{table*}

 \begin{figure*}[t]
  \centering
  \includegraphics[width = \linewidth]{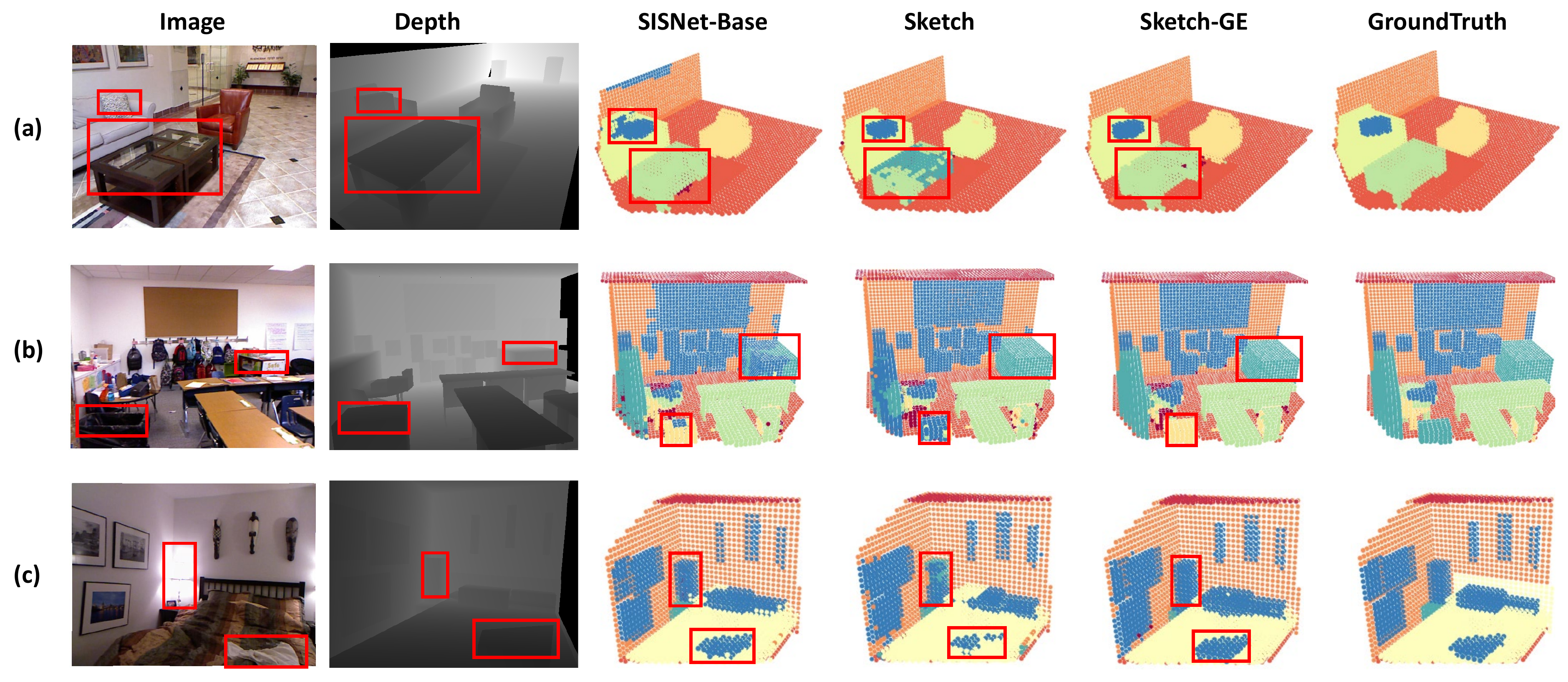}
  \caption{ Qualitative results on NYUCAD. All methods above are evaluated on both the observed and occluded voxels in the view frustum. We highlight some regions with {\color{red}\textit{red}} bounding box, which clearly show the effectiveness of our proposed block.} 
  \label{fig:case_ssc}
\end{figure*}

\subsubsection{Model Design Analysis}
To examine the effectiveness of our designs, we conduct a detailed ablation study on the key components of AdaPoinTr. The results are summarized in
Table~\ref{tab:Ablation Study}. The baseline model A is the vanilla Transformer model for point cloud completion, which uses the encoder-decoder architecture with the standard Transformer blocks. In this model, we form the point proxies directly from the point cloud using a single-layer DGCNN model. We then add the query generator between the encoder and decoder (model B). We see the query generator improve the baseline by 0.34 in Chamfer distance. When using DGCNN to extract features from the input point cloud (model C), we observe a significant improvement to 8.69. By adding the geometric block to all the Transformer blocks (model D), we see the performance can be further improved, which clearly demonstrates the effectiveness of the geometric structures learned by the block. We find that only adding the geometric block to the first Transformer block in both encoder and decoder can lead to a slightly better performance (PoinTr), which indicates the role of the geometric block is to introduce the inductive bias and a single layer is sufficient while adding more blocks may result in over-fitting. {\color{black} We then adopt the adaptive query generation mechanism on PoinTr, which achieve an improvement about 1.28 (model F). By adding the dynamic selection, the model can be more flexible when dealing with diverse categories and situations, which brings 0.14 improvement (model G). Finally, by introducing the auxiliary denoising task, our AdaPoinTr achieves the-state-of-art performance.}

\subsubsection{Complexity Analysis}
Our method achieves the best performance on both our newly proposed diverse benchmarks and the existing benchmarks. We provide the detailed complexity analysis of our method in Table~\ref{tab:FLOPs}. We report theoretical computation cost (FLOPs) and Throughput (Point Clouds Per Second) of our method and other methods. We also provide the results on {\color{black}ShapeNet-55/Projected-ShapeNet-55, unseen categories in ShapeNet-34/Projected-ShapeNet-34} and PCN benchmark as references. We observe that many recent advanced methods sacrifice efficiency in pursuit of higher performance while our method achieves the best performance on those benchmarks with less FLOPs and can process more than 60 point clouds within one second (using a batch size of 1). We argue that Throughputs should be paid more attention since it determines whether the model can be applied in some real-time situations.

\subsubsection{Qualitative Results}
In Fig.~\ref{fig:case}, we show some completion results for all methods and find our method performs better. For example, the input data in (a) nearly lose all the geometric information and can be hardly recognized as an airplane. In this case, other methods can only roughly complete the shape with unsatisfactory geometry details (clear contour of the wings), while our method can still complete the point cloud with higher fidelity. These results show our method has a stronger ability to recover details and is more robust to various incomplete patterns.

{\color{black}
\subsection{Semantic Scene Completion}
\label{sec:exp2}
\subsubsection{Benckmarks for Semantic Scene Completion}
In our experiments, we evaluate our method on two real-world datasets. These two datasets are the popular NYU Depth V2~\cite{silberman2012indoor} (denoted as NYUV2 in the following part) and NYUCAD~\cite{firman2016structured}. They both consist of 1449 indoor scenes and we follow the previous work to divide the datasets into training and test split, containing 795 and 654 scenes respectively. We follow the same experiment setting as~\cite{song2017semantic}.

\subsubsection{Evaluation Metric}
In our experiments, we follow~\cite{song2017semantic} to consider two sets of evaluation metrics: evaluation metric for scene completion (SC) and for semantic scene completion (SSC). Following ~\cite{firman2016structured}, we do not consider the voxels outside the view or the room while evaluating.

\paragrapha{5pt}{The evaluation metrics for SC.} We focus on the occupancy prediction results for each voxel in the scene. We follow the previous work to perform a binary prediction, \ie, empty or non-empty. We use \textit{recall}, \textit{precision} and voxel-wise intersection over union (\textit{IoU}) as the evaluation metrics to evaluate the performance on occluded voxels in the view frustum.

\paragrapha{5pt}{The evaluation metrics for SSC.} We focus on the semantic understanding of the entire scene. We regard all the empty voxels as one additional category and evaluate the intersection over union (\textit{IoU}) for each category on both the observed and occluded voxels in the view frustum.

\subsubsection{Results on NYUV2 and NYUCAD}

\paragrapha{5pt}{Results on NYUV2 and NYUCAD.}
On NYUV2 and NYUCAD, we implement the proposed block to the existing SOTA model, donated as DDRNet-GE and Sketch-GE. As shown in the Tabel~\ref{tab:NYUV2}, we compare our methods with other methods under SC and SSC tasks. The proposed block improves the performance of DDRNet~\cite{li2019rgbd} and Sketch and establishes a new SOTA for the end-to-end SSC model. Furthermore, We convert the depth map of a scene into a point cloud, and use PoinTr with Point-to-Voxel Translation (see Sec.~\ref{sec:p2v}) to treat it as a larger-scale completion task, we show the result in the table.

\subsubsection{Qualitative Results}
\label{sec:sscquan}
{\color{black}
In Fig.~\ref{fig:case_ssc}, we compare our method with other end-to-end methods for SSC and show some completion results from SISNet-Base~\cite{cai2021semantic} and Sketch~\cite{chen20203d}. By comparing Sketch-GE and Sketch~\cite{chen20203d}, we can find that our proposed block can enhance the geometric information for instances in the scenes. 
For example, our block helps the model to correctly recognize the table in (a), chair in (b) and object in (c). 
The good performance for our method is largely based on the geometrical knowledge and point-wise interactions learned by Transformers. 
By leveraging the geometric relations with the proposed block, models are encouraged to be aware of the existence of objects in scenes and effectively propagate and integrate the information from objects and scenes.

}

\section{Conclusion}
\label{sec:con}
In this paper, we have proposed a new architecture, PoinTr, to convert the point cloud completion task into a set-to-set translation task. With several technical innovations, we successfully applied the Transformer model to this task and achieved state-of-the-art performance. Moreover, we proposed {\color{black}four} more challenging benchmarks for more diverse {\color{black}object} point cloud completion.   {\color{black} We also verify that PoinTr is helpful to scene-level tasks. 
We expect introductions of PoinTr can provide some inspiration for the researchers in this area and we think }extending our Transformer architecture to more 3D tasks can an interesting future direction.

\vspace{-5pt}

\section*{Acknowledgement}

This work was supported in part by the National Key Research and Development Program of China under Grant 2017YFA0700802, in part by the National Natural Science Foundation of China under Grant 62125603, Grant 61822603, Grant U1813218, Grant U1713214, in part by Beijing Academy of Artificial Intelligence (BAAI), and in part by a grant from the Institute for Guo Qiang, Tsinghua University.

\vspace{-5pt}
\bibliographystyle{IEEEtranS}
\bibliography{ref}


\appendices

\section{Technical Details on Transformers}

\noindent \textbf{Encoder-Decoder Architecture. } The overall architecture of the Transformer encoder-decoder networks is illustrated in Fig.~\ref{fig:trans}. The point proxies are passed through the Transformer encoder with $N$ multi-head self-attention layers and feed-forward network layers. Then, the decoder receives the generated query embeddings and encoder memory, and produces the final set of predicted point proxies that represents the missing part of the point cloud through $N$ multi-head self-attention layers, decoder-encoder attention layers and feed-forward network layers. We set $N$ to 6 in all our experiments following the common practice~\cite{Transformer}.

\vspace{5pt}\noindent \textbf{Multi-head Attention. } Multi-head attention mechanism allows the network to jointly attend to information from different representation subspaces at different positions~\cite{Transformer}. Specifically, given the input values $V$, keys $K$ and queries $Q$, the multi-head attention is computed by:
\begin{equation}
    \begin{split}
        \text{MultiHead}(Q, K, V) = W^O \text{Concat}(\text{head}_1, ..., \text{head}_h), \nonumber
    \end{split}
\end{equation}
where $W^O$ the weights of the output linear layer and each head feature can be obtained by:
\begin{equation}
    \begin{split}
        \text{head}_i = \text{softmax}(\frac{QW_i^Q (KW_i^K)^T}{\sqrt{d_k}}) VW_i^V \nonumber
    \end{split}
\end{equation}
where $W_i^Q$, $W_i^K$, and $W_i^V$ are the linear layers that project the inputs to different subspaces and $d_k$ is the dimension of the input features. 

\vspace{5pt} \noindent \textbf{Feed-forward network (FFN). } Following~\cite{Transformer}, we use two linear layers with ReLU
activations and dropout as the feed-forward network.

 \begin{figure}[!ht]
  \centering
  \includegraphics[width = 0.8\linewidth]{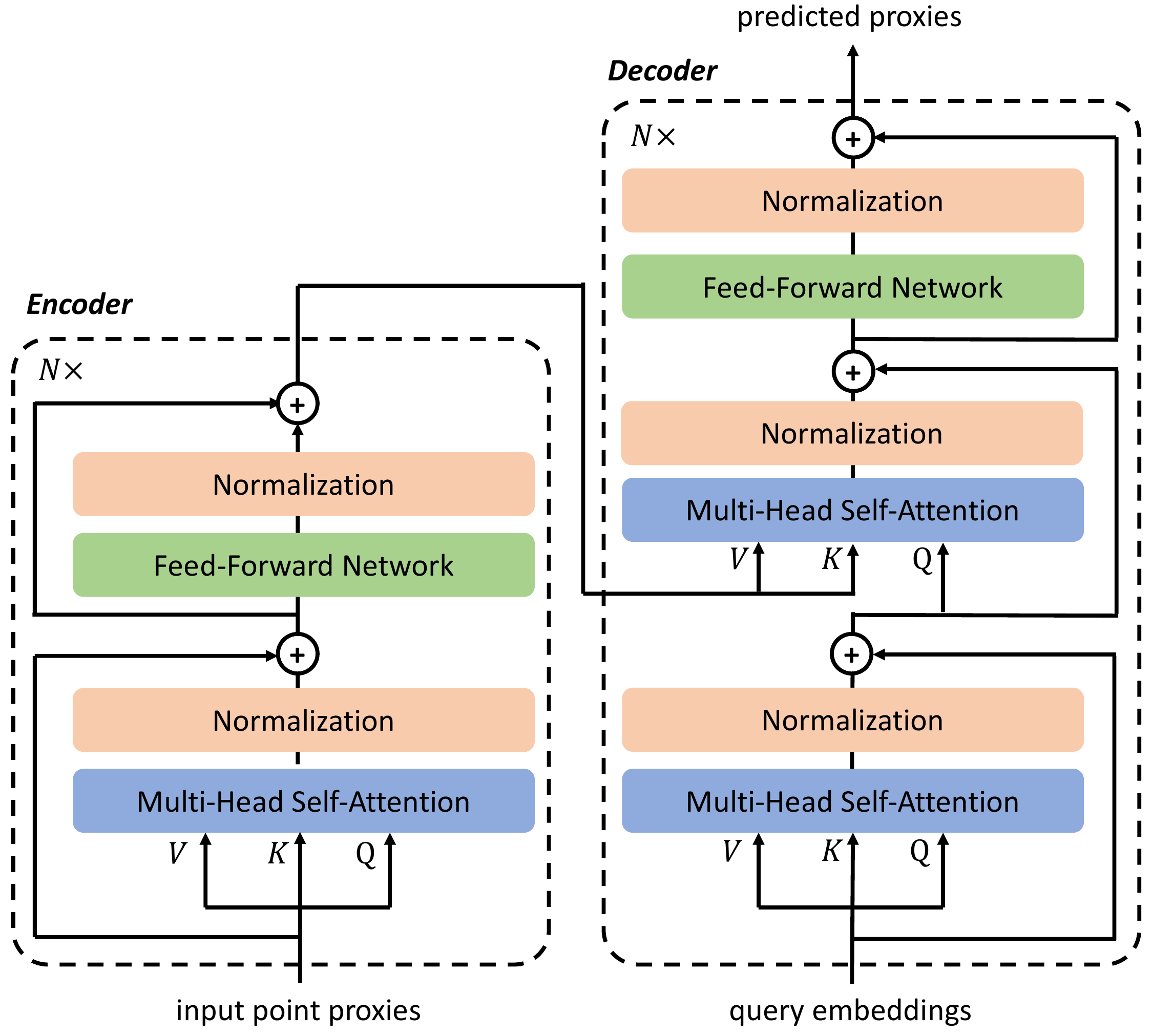}
  \caption{\small The overall architecture of the Transformer encoder-decoder networks.}
  \label{fig:trans}
\end{figure}

\section{Implementation Details}  
\paragrapha{5pt}{Point Proxies:} In our experiments, we convert the point cloud into a set of point proxies and employ a lightweight DGCNN~\cite{wang2019DGCNN} model to extract the point proxy features. To reduce the computational cost, we hierarchically downsample the original input point cloud to $N$ center points and use several DGCNN layers to capture local geometric relationships. The detailed network architecture is: $\texttt{Linear}(C_{in}=3, C_{out}=8)$ $\rightarrow$ $\texttt{DGCNN}(C_{in}=8, C_{out}=32, K=8, N_{out}=2048)$ $\rightarrow$ $\texttt{DGCNN}(C_{in}=32, C_{out}=64, K=8, N_{out}=512)$ $\rightarrow$ $\texttt{DGCNN}(C_{in}=64, C_{out}=64, K=8, N_{out}=512)$ $\rightarrow$ $\texttt{DGCNN}(C_{in}=64, C_{out}=128, K=8, N_{out}=N)$, where $C_{in}$ and $C_{out}$ are the numbers of channels of input and output features, $N_{out}$ is the number of points after FPS. We set $N$ to 256 for experiments point cloud completion and semantic scene completion, respectively.

\paragrapha{5pt}{Object Point Cloud Completion:} We utilize AdamW optimizer~\cite{loshchilov2018fixing} to train the network with initial learning rate as 0.0001 and weight decay as 0.0005. In all of our experiments, we set the depth of the encoder and decoder in our Transformer to 6 and 8 and set $k$ of kNN operation to 16 and 8 for the DGCNN feature extractor and the geometry-aware block respectively. We use 6 head attention for all Transformer blocks and set their hidden dimensions to 384. On the PCN dataset, the network takes 2048 points as inputs and is required to complete the other 14336 points. We set the batch size to 48 and train the model for 300 epochs with the continuous learning rate decay of 0.9 for every 20 epochs. We set $N$ to 256 and $M$ to 512 (including 256 queries from input point proxies). We add 64 \textit{Denoise Queries} during training phase. On ShapeNet-55/34 and Projected-ShapeNet-55/34, the model takes 2048 points as inputs and is required to complete the other 6144 points. We set the batch size to 64 for ShapeNet-55/34 and Projected-ShapeNet-55/34. We train the model for 300 epochs with the continuous learning rate decay of 0.76 for every 20 epochs. The number of \textit{Dynamic Queries} $M$ is set to 256 and the number of \textit{Denoise Queries} is set to 64. During the inference, we skip the denoise process and only focus on completing point cloud from the generated queries.

\paragrapha{5pt}{Semantic Scene Completion:} We follow the previous work to utilize SGD optimizer~\cite{ruder2016overview} during the training phase. The learning rate and the weight decay are set to 0.1 and 0.0005, respectively. We adjust the learning rate from 0.1 to 0.00001 with a cosine schedule and set the batch size to 4. In all of our experiments, we set the depth of the encoder and decoder in our Transformer to 3 and 3, and $k$ of kNN operation to 16 and 8 for the DGCNN feature extractor and the geometry-aware block respectively. We use 6 head attention for all Transformer blocks and set their hidden dimensions to 384. The number of \textit{Voxel Queries} for the Transformer decoder is set to 75.

\section{Qualitative Results} 
We provide more qualitative results for PoinTr on point cloud completion and semantic scene completion in this part.

\paragrapha{5pt}{Predicted Centers}
We visualize the local center prediction results on ShapeNet-55. We adopt a coarse-to-fine strategy to recover the point cloud. Our method starts with the prediction of local centers, then we can obtain the final results by adding the points around the centers. As shown in Fig.~\ref{fig:proxy}, Line (a) shows the input partial point cloud and the predicted point centers. Line (b) is the predicted point cloud. We see the predicted point proxies can successfully represent the overall structure of the point cloud and the details then are added in the final predictions.

 \begin{figure}[t]
  \centering
  \includegraphics[width = \linewidth]{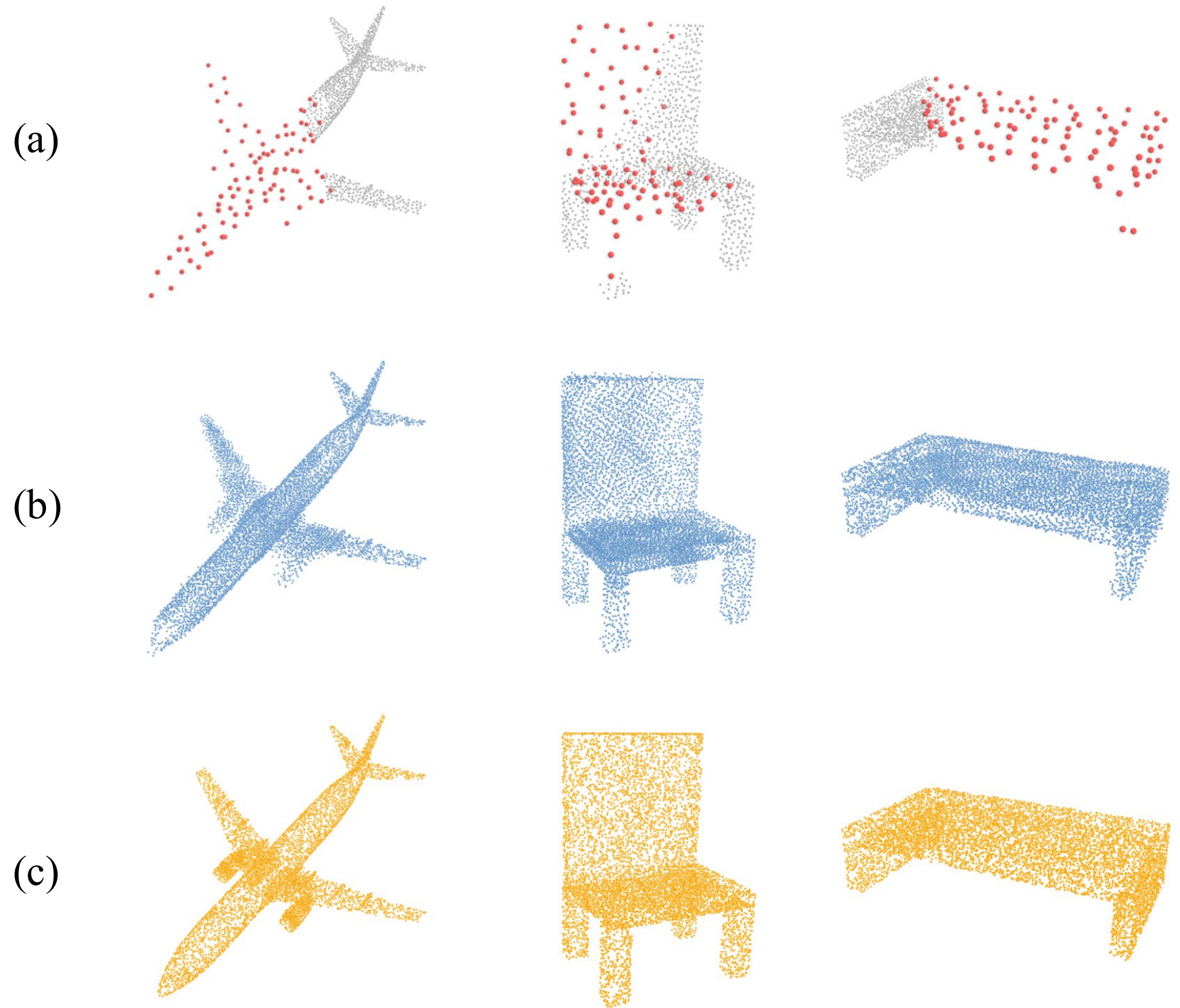}
  \caption{\small Visualization of predicted points proxies. In Line (a), we show the input partial point clouds and the predicted centers. Based on predicted point proxies, we can easily predict the accurate point centers and then complete the point clouds, as shown in Line (b). We show the ground-truth point cloud in Line (c) for comparisons.}
  \label{fig:proxy}
\end{figure}

\paragrapha{5pt}{Point Cloud Completion:} In Fig.~\ref{fig:supp_case2}, we show more completion results on ShapeNet-55.  We see that our PoinTr has a stronger ability to recover details and is more robust to
various incomplete patterns.

\paragrapha{5pt}{Semantic Scene Completion:} In Fig.~\ref{fig:ssc_supp_case2}, we show more completion results on NYUCAD~\cite{firman2016structured}. We see our Geometry-Enhanced block can help to keep more geometry information and guide the final prediction.

\section{More Experimental Results }

\vspace{5pt} \noindent \textbf{Ablation on the number of point proxies and the number of queries: } 
We perform the ablation studies about the number of input point proxies $N$ and the number of generated queries $q$ on widely-used PCN benchmark as below.
\begin{table}[!h]
    \centering \small
\newcolumntype{g}{>{\columncolor{Gray}}l}
  \begin{tabular}{l|ccg}\toprule
     CD-$\ell_1$ & N=128 & N=256 & N=512 \\ \midrule
    $q=128$  & 6.82  & 6.78  & 6.75  \\ 
    $q=256$  & 6.68  & 6.65  & 6.63  \\ 
    \rowcolor{Gray}$q=512$  & 6.57  & \textbf{6.51}  & 6.53  \\  \midrule
\end{tabular}
\end{table}
We can see that, increasing the number of queries $q$ or the number of input point proxies $N$ can both bring improvement. However, we can not increase the number infinitely due to the quadratic complexity of attention mechanism, which will bring an unbearable computational cost. Moreover, we observe a saturation trend when $N > 256$, while the situation is different for $q$, which is because there are only 2048 points in the input point cloud.

\noindent \textbf{Ablation on formulation of point proxies: } As for the presentation of point proxies, we have conducted ablation studies to investigate the gathering operation and the feature extractor.
\begin{table}[!h]
    \centering \small
\newcolumntype{g}{>{\columncolor{Gray}}l}
  \begin{tabular}{l|cc}\toprule
     & CD-$\ell_1$   \\ \midrule
    AdaPoinTr  & 6.51   \\ 
    - Abandoning Positional Embedding  & 6.89 ({\color{green} -0.48})    \\
    - Replace mini-DGCNN with PointNet & 7.12 ({\color{green} -0.23})    \\ 
\midrule
\end{tabular}
\end{table}
 We build the point proxies as $F_i = F'_i + \varphi(p_i)$, where $F'_i$ is the local structure feature around the point $p_i$ and $\varphi(p_i)$ is the positional embedding. We first abandon the positional embedding and build the point proxies as $F_i = F'_i$, the performance drops 0.48. Then, we replace the feature extractor from mini-DGCNN to PointNet, and the performance drops 0.23.

\section{Additional Description to KITTI Metric}
We adopt the same definition for those metric on KITTI benchmark as PCN~\cite{PCN}

\noindent \textbf{MMD: } Minimal Matching Distance (MMD), which is the Chamfer Distance (CD) between the output and the car point cloud from PCN~\cite{PCN} that is closest to the output point cloud in terms of CD. This measures how much the output resembles a typical car;

\noindent \textbf{Fidelity: } Fidelity is the average distance from each point in the input to its nearest neighbour in the output. This measures how well the input is preserved;

\section{Detailed Experimental Results }

\vspace{5pt} \noindent \textbf{Detailed results on ShapeNet-34: } 
In Table~\ref{Table:ShapeNet-21}, we report the detailed results of FoldingNet~\cite{FoldingNet}, PCN~\cite{PCN}, TopNet~\cite{TopNet}, PFNet~\cite{PFNet}, GRNet~\cite{GRNet}, SnowflakeNet~\cite{xiang2021snowflakenet} and the proposed method for the novel objects from 21 categories in ShapeNet-34. Each row in the table stands for a category of objects. We test each method under the three settings: simple, moderate and hard and use CD-$\ell_2$ as the evaluation metric. 

\vspace{5pt} \noindent \textbf{Detailed results on Projected-ShapeNet-34: } 
In Table~\ref{Table:PShapeNet-34}, we report the detailed results for the novel objects from 21 categories in Projected-ShapeNet-34. Each row in the table stands for a category of objects. We report CD-$\ell_1$ and F-score@1\% for each method.

\vspace{5pt} \noindent \textbf{Detailed results on ShapeNet-55:} In Table~\ref{Table:ShapeNet-55}, we report the detailed results of each method on ShapeNet-55. Each row in the table stands for a category of objects. We test each method under three settings: simple, moderate and hard and use CD-$\ell_2$ as the evaluation metric. 

\vspace{5pt} \noindent \textbf{Detailed results on Projected-ShapeNet-55:} In Table~\ref{Table:PShapeNet-55}, we report the detailed results of each method on Projected-ShapeNet-55. Each row in the table stands for a category of objects. We report CD-$\ell_1$ and F-score@1\% for each method.

 \begin{figure*}
  \centering
  \includegraphics[width = \linewidth]{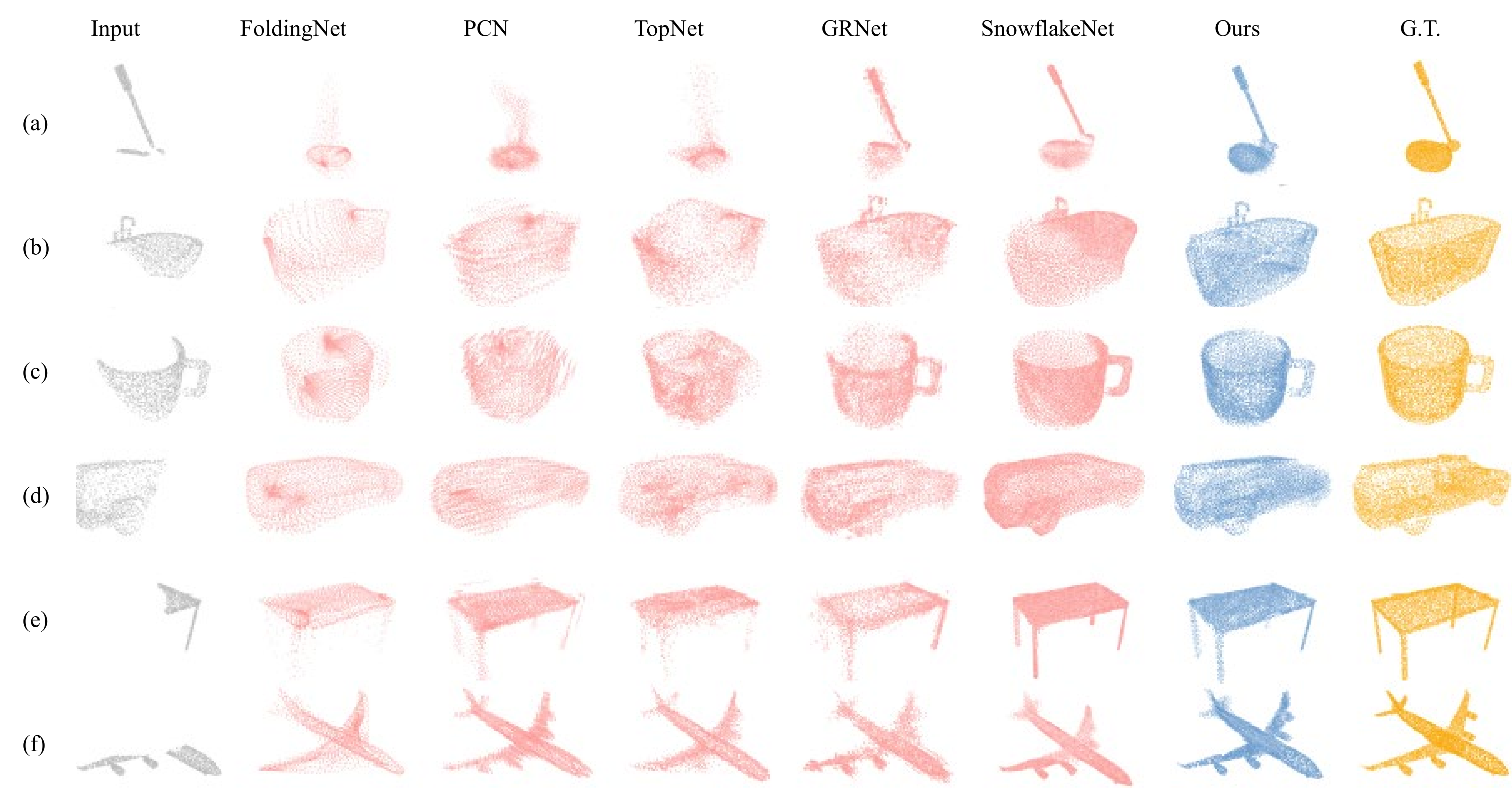}
  \caption{\small More qualitative results on ShapeNet-55.}
  \label{fig:supp_case2}
\end{figure*}

 \begin{figure*}
  \centering
  \includegraphics[width = \linewidth]{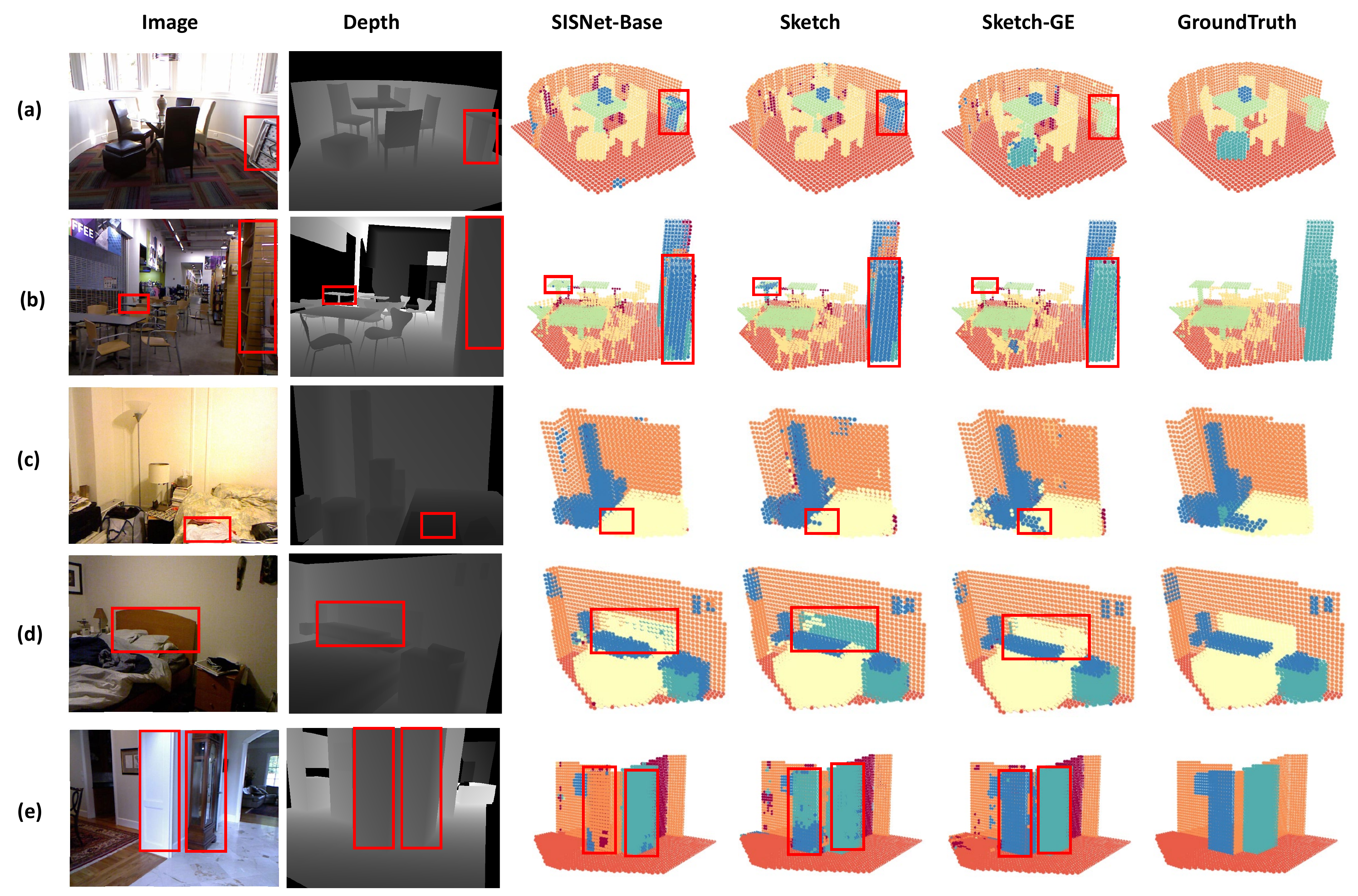}
  \caption{\small More qualitative results on NYUCAD.}
  \label{fig:ssc_supp_case2}
\end{figure*}

\begin{table*}
\caption{
Detailed results under CD-$\ell_2$ (multiplied by 1000) for the novel objects on ShapeNet-34. \textit{S.}, \textit{M.} and \textit{H.} stand for the simple, moderate and hard settings.}
\vspace{4pt}
\label{Table:ShapeNet-21} \centering
\setlength{\tabcolsep}{3pt}{
\adjustbox{width=\linewidth}{
\begin{tabular}{@{\hskip 5pt}>{\columncolor{white}[5pt][\tabcolsep]}l | c c c| c c c| c c c| c c c| c c c| c c c| c c>{\columncolor{white}[\tabcolsep][5pt]}c@{\hskip 5pt}}

\toprule
\multirow{2}{*}{}& \multicolumn{3}{c|}{FoldingNet~\cite{FoldingNet}}&\multicolumn{3}{c|}{PCN~\cite{PCN}}&\multicolumn{3}{c|}{TopNet~\cite{TopNet}}&\multicolumn{3}{c|}{PFNet~\cite{PFNet}}&\multicolumn{3}{c|}{GRNet~\cite{GRNet}}&\multicolumn{3}{c|}{SnowflakeNet~\cite{xiang2021snowflakenet}}&\multicolumn{3}{c}{AdaPoinTr}\\
\cmidrule{2-22}
&S.&M.&H.&S.&M.&H.&S.&M.&H.&S.&M.&H.&S.&M.&H.&S.&M.&H.&S.&M.&H.\\
\midrule
bag& 2.15 & 2.27 & 3.99 & 2.48 & 2.46 & 3.94 & 2.08 & 1.95 & 4.36 & 3.88 & 4.42 & 9.67 & 1.47 & 1.88 & 3.45 & 0.67 & 1.08 & 1.82 & \textbf{0.54}& \textbf{0.75}& \textbf{1.23}\\
basket& 2.37 & 2.2 & 4.87 & 2.79 & 2.51 & 4.78 & 2.46 & 2.11 & 5.18 & 4.47 & 4.55 & 14.46 & 1.78 & 1.94 & 4.18 & 0.78 & 1.16 & 2.48 &\textbf{0.67}& \textbf{0.82}& \textbf{1.57}\\
birdhouse& 3.27 & 3.15 & 5.62 & 3.53 & 3.47 & 5.31 & 3.17 & 2.97 & 5.89 & 3.9 & 4.65 & 9.88 & 1.89 & 2.34 & 5.16 & 0.95 & 1.46 & 2.78 & \textbf{0.75}& \textbf{1.06}& \textbf{1.96}\\
bowl& 2.61 & 2.3 & 4.55 & 2.66 & 2.35 & 3.97 & 2.46 & 2.16 & 4.84 & 4.35 & 5.0 & 14.59 & 1.77 & 1.97 & 3.9 & 0.77 & 1.15 & 2.03 & \textbf{0.66}& \textbf{0.77}& \textbf{1.13}\\
camera& 4.4 & 4.78 & 7.85 & 4.84 & 5.3 & 8.03 & 4.24 & 4.43 & 8.11 & 6.78 & 8.04 & 13.91 & 2.31 & 3.38 & 7.2 & 1.27 & 2.30 & 4.31 & \textbf{0.85}& \textbf{1.56}& \textbf{3.33}\\
can& 1.95 & 1.73 & 5.86 & 1.95 & 1.89 & 5.21 & 2.02 & 1.7 & 5.82 & 2.95 & 3.47 & 23.02 & 1.53 & 1.8 & 3.08 &\textbf{ 0.62} & 0.94 & 1.73 & 0.64& \textbf{0.88}& \textbf{1.48}\\
cap& 6.07 & 5.98 & 11.49 & 7.21 & 7.14 & 10.94 & 4.68 & 4.23 & 9.17 & 14.11 & 14.86 & 28.23 & 3.29 & 4.87 & 13.02 &1.29  & 3.10 &7.37&  \textbf{0.47}& \textbf{0.98}& \textbf{3.70}\\
keyboard& 0.98 & 0.96 & 1.35 & 1.07 & 1.0 & 1.23 & 0.79 & 0.77 & 1.55 & 1.13 & 1.16 & 2.58 & 0.73 & 0.77 & 1.11 & 0.38 & 0.46 & 0.63 & \textbf{0.32}& \textbf{0.36}& \textbf{0.45}\\
dishwasher& 2.09 & 1.8 & 4.55 & 2.45 & 2.09 & 3.53 & 2.51 & 1.77 & 4.72 & 3.44 & 3.78 & 9.31 & 1.79 & 1.7 & 3.27 & \textbf{0.67} & 0.89 & 1.66 &  0.73& \textbf{0.83}& \textbf{1.28}\\
earphone& 6.86 & 6.96 & 12.77 & 7.88 & 6.59 & 16.53 & 5.33 & 4.83 & 11.67 & 20.31 & 23.21 & 39.49 & 4.29 & 4.16 & 10.3 & 2.29 & 4.01 & 9.61 & \textbf{1.17}& \textbf{2.24}& \textbf{7.50}\\
helmet& 4.86 & 5.04 & 8.86 & 6.15 & 6.41 & 9.16 & 4.89 & 4.86 & 8.73 & 8.78 & 10.07 & 21.2 & 3.06 & 4.38 & 10.27 & 1.74 & 2.96  & 5.87 & \textbf{1.05}& \textbf{2.12}& \textbf{4.74}\\
mailbox& 2.2 & 2.29 & 4.49 & 2.74 & 2.68 & 4.31 & 2.35 & 2.2 & 4.91 & 5.2 & 5.33 & 10.94 & 1.52 & 1.9 & 4.33 &0.83 & 1.36 & 2.43 & \textbf{0.43}& \textbf{0.78}& \textbf{1.93}\\
microphone& 2.92 & 3.27 & 8.54 & 4.36 & 4.65 & 8.46 & 3.03 & 3.2 & 7.15 & 6.39 & 7.99 & 19.41 & 2.29 & 3.23 & 8.41 & 1.63 & 2.40 & 5.18 & \textbf{0.69}& \textbf{1.54}& \textbf{3.95}\\
microwaves& 2.29 & 2.12 & 5.17 & 2.59 & 2.35 & 4.47 & 2.67 & 2.12 & 5.41 & 3.89 & 4.08 & 9.01 & 1.74 & 1.81 & 3.82 &\textbf{ 0.72} & 0.96 & 1.78 & 0.74& \textbf{0.85}& \textbf{1.51}\\
pillow& 2.07 & 2.11 & 3.73 & 2.09 & 2.16 & 3.54 & 2.08 & 2.05 & 4.01 & 4.15 & 4.29 & 12.01 & 1.43 & 1.69 & 3.43 &  0.55& 0.86 & 1.75 &\textbf{0.45}& \textbf{0.56}& \textbf{1.06}\\
printer& 3.02 & 3.23 & 5.53 & 3.28 & 3.6 & 5.56 & 2.9 & 2.96 & 6.07 & 5.38 & 5.94 & 10.29 & 1.82 & 2.41 & 5.09 &0.87 & 1.65 & 2.72 & \textbf{0.69}& \textbf{1.08}& \textbf{2.09}\\
remote& 0.89 & 0.92 & 1.85 & 0.95 & 1.08 & 1.58 & 0.89 & 0.89 & 2.28 & 1.51 & 1.75 & 6.0 & 0.82 & 1.02 & 1.29 & 0.32 & 0.47 & 0.67 & \textbf{0.29}& \textbf{0.38}& \textbf{0.48}\\
rocket& 1.28 & 1.09 & 2.0 & 1.39 & 1.22 & 2.01 & 1.14 & 0.96 & 2.03 & 1.84 & 1.51 & 4.01 & 0.97 & 0.79 & 1.6  & 0.36 & 0.57 & 1.05 &  \textbf{0.25}& \textbf{0.50}& \textbf{0.95}\\
skateboard& 1.53 & 1.42 & 1.99 & 1.97 & 1.78 & 2.45 & 1.23 & 1.2 & 2.01 & 2.43 & 2.53 & 4.25 & 0.93 & 1.07 & 1.83 & 0.48& 0.77 & 1.21 & \textbf{0.29}& \textbf{0.49}& \textbf{0.73}\\
tower& 2.25 & 2.25 & 4.74 & 2.37 & 2.4 & 4.35 & 2.2 & 2.17 & 5.47 & 3.38 & 4.15 & 13.11 & 1.35 & 1.8 & 3.85 & 0.67 & 1.12 & 2.20 &\textbf{0.48}& \textbf{0.83}& \textbf{1.65}\\
washer& 2.58 & 2.34 & 5.5 & 2.77 & 2.52 & 4.64 & 2.63 & 2.14 & 6.57 & 4.53 & 4.27 & 9.23 & 1.83 & 1.97 & 5.28 & 0.75  & 1.07 &2.23 &\textbf{0.70}& \textbf{0.86}& \textbf{1.67}\\
\midrule
mean& 2.79 & 2.77 & 5.49 & 3.22 & 3.13 & 5.43 & 2.65 & 2.46 & 5.52 & 5.37 & 5.95 & 13.55 & 1.84 & 2.23 & 4.95 & 0.88 & 1.46 & 2.92 & \textbf{0.61}& \textbf{0.96}& \textbf{2.11}\\

\bottomrule
\end{tabular}}}
\end{table*}

\begin{table*}
\caption{
Detailed results CD-$\ell_1$ (multiplied by 1000) and F-Score@1\% on Projected-ShapeNet-34.}
\label{Table:PShapeNet-34} 
\centering
\setlength{\tabcolsep}{8pt}{
\adjustbox{width=\linewidth}{
\begin{tabular}{@{\hskip 5pt}>{\columncolor{white}[5pt][\tabcolsep]}l | cc|cc|cc|cc|cc|c>{\columncolor{white}[\tabcolsep][5pt]}c@{\hskip 5pt}}
\toprule
\multirow{2}{*}{ }&\multicolumn{2}{c|}{PCN~\cite{PCN}}&\multicolumn{2}{c|}{TopNet~\cite{TopNet}}&\multicolumn{2}{c|}{GRNet~\cite{GRNet}}&\multicolumn{2}{c|}{SnowflakeNet~\cite{GRNet}}&\multicolumn{2}{c|}{PoinTr~\cite{xiang2021snowflakenet}} &\multicolumn{2}{c}{AdaPoinTr}\\
\cmidrule{2-13} 
& CD & F-Score & CD & F-Score & CD & F-Score & CD & F-Score & CD & F-Score & CD & F-Score\\
\midrule
bag& 20.61& 0.269& 15.96& 0.330& 15.25& 0.407& 12.90& 0.518& 12.87& 0.520& \textbf{11.87}& \textbf{0.59}\\
basket& 19.54& 0.278& 14.70& 0.373& 15.12& 0.380& 12.55& 0.517& 11.98& 0.550& \textbf{11.22}& \textbf{0.62}\\
birdhouse& 20.75& 0.259& 18.06& 0.246& 16.21& 0.373& 14.55& 0.450& 13.83& 0.459& \textbf{12.78}& \textbf{0.54}\\
bowl& 15.66& 0.406& 12.74& 0.470& 13.25& 0.478& 10.92& 0.622& 10.45& 0.663& \textbf{9.68}& \textbf{0.74}\\
camera& 24.58& 0.204& 19.40& 0.216& 16.70& 0.369& 14.48& 0.458& 14.22& 0.461& \textbf{13.22}& \textbf{0.53}\\
can& 20.16& 0.253& 15.45& 0.356& 15.61& 0.363& 12.98& 0.496& 12.24& 0.546& \textbf{11.02}& \textbf{0.64}\\
cap& 25.86& 0.231& 17.90& 0.314& 16.01& 0.467& 13.54& 0.601& 12.75& 0.599& \textbf{11.30}& \textbf{0.71}\\
dishwasher& 23.16& 0.208& 17.19& 0.270& 17.39& 0.290& 15.21& 0.416& 14.40& 0.421& \textbf{13.39}& \textbf{0.50}\\
earphone& 24.82& 0.287& 19.34& 0.297& 15.00& 0.485& 15.19& 0.532& 14.23& 0.505& \textbf{12.30}& \textbf{0.61}\\
helmet& 23.46& 0.252& 19.04& 0.261& 16.71& 0.396& 14.65& 0.490& 14.27& 0.497& \textbf{13.50}& \textbf{0.56}\\
keyboard& 14.59& 0.529& 10.48& 0.589& 10.12& 0.621& 8.78& 0.710& 8.87& 0.710& \textbf{8.00}& \textbf{0.77}\\
mailbox& 22.34& 0.292& 16.93& 0.316& 14.30& 0.469& 12.53& 0.580& 12.00& 0.583& \textbf{10.51}& \textbf{0.70}\\
microphone& 19.05& 0.375& 14.34& 0.387& 11.39& 0.577& 10.10& 0.679& 9.34& 0.672& \textbf{7.96}& \textbf{0.79}\\
microwaves& 39.08& 0.148& 19.41& 0.231& 19.59& 0.270& 15.41& 0.396& 15.76& 0.385& \textbf{15.39}& \textbf{0.47}\\
pillow& 23.45& 0.230& 17.55& 0.336& 18.64& 0.373& 15.35& 0.494& 14.82& 0.502& \textbf{14.19}& \textbf{0.58}\\
printer& 27.89& 0.194& 18.86& 0.244& 18.00& 0.331& 14.74& 0.442& 14.94& 0.437& \textbf{13.72}& \textbf{0.52}\\
remote& 13.73& 0.463& 11.36& 0.547& 12.19& 0.513& 9.92& 0.686& 9.54& 0.701& \textbf{8.09}& \textbf{0.80}\\
rocket& 11.27& 0.584& 10.07& 0.621& 9.57& 0.656& 7.90& 0.774& 7.97& 0.719& \textbf{6.84}& \textbf{0.84}\\
skateboard& 17.27& 0.419& 12.59& 0.517& 10.60& 0.666& 9.58& 0.740& 8.98& 0.737& \textbf{8.34}& \textbf{0.80}\\
tower& 17.42& 0.384& 15.39& 0.372& 14.52& 0.473& 12.50& 0.565& 12.18& 0.582& \textbf{10.95}& \textbf{0.67}\\
washer& 25.62& 0.176& 18.71& 0.234& 19.45& 0.270& 15.44& 0.401& 15.43& 0.406& \textbf{14.49}& \textbf{0.50}\\
\midrule
mean & 21.44& 0.307& 15.98& 0.358& 15.03& 0.439& 12.82& 0.551& 12.43& 0.555& \textbf{11.37}& \textbf{0.64}\\
\bottomrule
\end{tabular}}}
\vspace{20pt}
\end{table*}

\begin{table*}
\caption{
Detailed results under CD-$\ell_2$ (multiplied by 1000) on ShapeNet-55. \textit{S.}, \textit{M.} and \textit{H.} stand for the simple, moderate and hard settings.}
\label{Table:ShapeNet-55} 
\centering
\setlength{\tabcolsep}{3pt}{
\adjustbox{width=\linewidth}{
\begin{tabular}{@{\hskip 5pt}>{\columncolor{white}[5pt][\tabcolsep]}l | c c c| c c c| c c c| c c c| c c c| c c c| c c>{\columncolor{white}[\tabcolsep][5pt]}c@{\hskip 5pt}}

\toprule
\multirow{2}{*}{}& \multicolumn{3}{c|}{FoldingNet~\cite{FoldingNet}}&\multicolumn{3}{c|}{PCN~\cite{PCN}}&\multicolumn{3}{c|}{TopNet~\cite{TopNet}}&\multicolumn{3}{c|}{PFNet~\cite{PFNet}}&\multicolumn{3}{c|}{GRNet~\cite{GRNet}}&\multicolumn{3}{c|}{SnowflakeNet~\cite{xiang2021snowflakenet}} &\multicolumn{3}{c}{AdaPoinTr}\\
\cmidrule{2-22}
&S.&M.&H.&S.&M.&H.&S.&M.&H.&S.&M.&H.&S.&M.&H.&S.&M.&H.&S.&M.&H.\\
\midrule
airplane& 1.36 & 1.28 & 1.7 & 0.9 & 0.89 & 1.32 & 1.02 & 0.99 & 1.48 & 1.35 & 1.44 & 2.69 & 0.87 & 0.87 & 1.27 & 0.36 & 0.50 & 0.76 & \textbf{0.22}& \textbf{0.30}& \textbf{0.49}\\
trash bin& 2.93 & 2.9 & 5.03 & 2.16 & 2.18 & 5.15 & 2.51 & 2.32 & 5.03 & 4.03 & 3.39 & 9.63 & 1.69 & 2.01 & 3.48 & 0.94 & 1.32 & 2.51& \textbf{0.75}& \textbf{0.98}& \textbf{1.61}\\
bag& 2.31 & 2.38 & 3.67 & 2.11 & 2.04 & 4.44 & 2.36 & 2.23 & 4.21 & 3.63 & 3.66 & 7.6 & 1.41 & 1.7 & 2.97 & 0.62 & 0.93 & 1.70& \textbf{0.45}& \textbf{0.62}& \textbf{1.00}\\
basket& 2.98 & 2.77 & 4.8 & 2.21 & 2.1 & 4.55 & 2.62 & 2.43 & 5.71 & 4.74 & 3.88 & 8.47 & 1.65 & 1.84 & 3.15 & 0.81 & 1.05 & 2.13& \textbf{0.68}& \textbf{0.78}& \textbf{1.28}\\
bathtub& 2.68 & 2.66 & 4.0 & 2.11 & 2.09 & 3.94 & 2.49 & 2.25 & 4.33 & 3.64 & 3.5 & 5.74 & 1.46 & 1.73 & 2.73 & 0.71 & 1.11 & 1.81& \textbf{0.53}& \textbf{0.72}& \textbf{1.18}\\
bed& 4.24 & 4.08 & 5.65 & 2.86 & 3.07 & 5.54 & 3.13 & 3.1 & 5.71 & 4.44 & 5.36 & 9.14 & 1.64 & 2.03 & 3.7 & 0.92 & 1.35 & 2.53& \textbf{0.63}& \textbf{0.86}& \textbf{1.64}\\
bench& 1.94 & 1.77 & 2.36 & 1.31 & 1.24 & 2.14 & 1.56 & 1.39 & 2.4 & 2.17 & 2.16 & 4.11 & 1.03 & 1.09 & 1.71 & 0.48 & 0.65 & 1.15& \textbf{0.31}& \textbf{0.37}& \textbf{0.64}\\
birdhouse& 4.06 & 4.18 & 5.88 & 3.29 & 3.53 & 6.69 & 3.73 & 3.98 & 6.8 & 3.96 & 5.0 & 9.66 & 1.87 & 2.4 & 4.71 & 1.06 & 1.69 & 3.04& \textbf{0.81}& \textbf{1.18}& \textbf{2.08}\\
bookshelf& 3.04 & 3.03 & 3.91 & 2.7 & 2.7 & 4.61 & 3.11 & 2.87 & 4.87 & 3.19 & 3.47 & 5.72 & 1.42 & 1.71 & 2.78 & 0.78 & 1.17 & 1.94& \textbf{0.61}& \textbf{0.79}& \textbf{1.38}\\
bottle& 1.7 & 1.91 & 4.02 & 1.25 & 1.43 & 4.61 & 1.56 & 1.66 & 4.02 & 2.37 & 2.89 & 10.03 & 1.05 & 1.44 & 2.67 & 0.44 & 0.80 & 1.54& \textbf{0.31}& \textbf{0.55}& \textbf{1.07}\\
bowl& 2.79 & 2.6 & 4.23 & 2.05 & 1.83 & 3.66 & 2.33 & 1.98 & 4.82 & 4.3 & 3.97 & 8.76 & 1.6 & 1.77 & 2.99 & 0.76 & 0.96 & 1.81& \textbf{0.60}& \textbf{0.63}& \textbf{0.92}\\
bus& 1.47 & 1.42 & 2.0 & 1.2 & 1.14 & 2.08 & 1.32 & 1.21 & 2.29 & 2.06 & 1.88 & 3.75 & 1.06 & 1.16 & 1.48 & 0.50 & 0.64 & 0.87& \textbf{0.43}& \textbf{0.52}& \textbf{0.65}\\
cabinet& 2.0 & 1.86 & 2.79 & 1.6 & 1.49 & 3.47 & 1.91 & 1.65 & 3.36 & 2.72 & 2.37 & 4.73 & 1.27 & 1.41 & 2.09 & 0.66 & 0.83 & 1.27& \textbf{0.59}& \textbf{0.65}& \textbf{0.92}\\
camera& 5.5 & 6.04 & 8.87 & 4.05 & 4.54 & 8.27 & 4.75 & 4.98 & 9.24 & 6.57 & 8.04 & 13.11 & 2.14 & 3.15 & 6.09 & 1.27 & 2.36 & 4.30& \textbf{0.80}& \textbf{1.55}& \textbf{3.15}\\
can& 2.84 & 2.68 & 5.71 & 2.02 & 2.28 & 6.48 & 2.67 & 2.4 & 5.5 & 5.65 & 4.05 & 16.29 & 1.58 & 2.11 & 3.81 & 0.63 & 1.23 & 2.23& \textbf{0.61}& \textbf{0.94}& \textbf{1.68}\\
cap& 4.1 & 4.04 & 5.87 & 1.82 & 1.76 & 4.2 & 3.0 & 2.69 & 5.59 & 10.92 & 9.04 & 20.3 & 1.17 & 1.37 & 3.05 & 0.55 & 0.94 & 1.98& \textbf{0.37}& \textbf{0.43}& \textbf{0.69}\\
car& 1.81 & 1.81 & 2.31 & 1.48 & 1.47 & 2.6 & 1.71 & 1.65 & 3.17 & 2.06 & 2.1 & 3.43 & 1.29 & 1.48 & 2.14 & 0.81 & 1.01 & 1.32& \textbf{0.65}& \textbf{0.79}& \textbf{1.00}\\
cellphon& 1.04 &  1.06& 1.87 & 0.8 & 0.79 & 1.71 & 1.01 & 0.96 & 1.8 & 1.25 & 1.37 & 3.65 & 0.82 & 0.91 & 1.18 & 0.37 & 0.49 & 0.71& \textbf{0.33}& \textbf{0.36}& \textbf{0.44}\\
chair& 2.37 & 2.46 & 3.62 & 1.7 & 1.81 & 3.34 & 1.97 & 2.04 & 3.59 & 2.94 & 3.48 & 6.34 & 1.24 & 1.56 & 2.73 & 0.62 & 0.94 & 1.79& \textbf{0.41}& \textbf{0.56}& \textbf{1.12}\\
clock& 2.56 & 2.41 & 3.46 & 2.1 & 2.01 & 3.98 & 2.48 & 2.16 & 4.03 & 3.15 & 3.27 & 6.03 & 1.46 & 1.66 & 2.67 & 0.74 & 1.00 & 1.65& \textbf{0.53}& \textbf{0.68}& \textbf{1.18}\\
keyboard& 1.21 & 1.18 & 1.32 & 0.82 & 0.82 & 1.04 & 0.88 & 0.83 & 1.15 & 0.83 & 1.06 & 1.97 & 0.74 & 0.81 & 1.09 & 0.36 & 0.45 & 0.63& \textbf{0.28}& \textbf{0.32}& \textbf{0.37}\\
dishwasher& 2.6 & 2.17 & 3.5 & 1.93 & 1.66 & 4.39 & 2.43 & 1.74 & 4.64 & 4.57 & 3.23 & 6.39 & 1.43 & 1.59 & 2.53 & 0.63 & 0.82 & 1.69& \textbf{0.60}& \textbf{0.66}& \textbf{1.16}\\
display& 2.15 & 2.24 & 3.25 & 1.56 & 1.66 & 3.26 & 1.84 & 1.85 & 3.48 & 2.27 & 2.83 & 5.52 & 1.13 & 1.38 & 2.29 & 0.57 & 0.90 & 1.57&\textbf{0.41}& \textbf{0.53}& \textbf{0.89}\\
earphone& 6.37 & 6.48 & 9.14 & 3.13 & 2.94 & 7.56 & 4.36 & 4.47 & 8.36 & 15.07 & 17.5 & 33.37 & 1.78 & 2.18 & 5.33 & 0.94 & 1.55 & 4.15&\textbf{0.60}& \textbf{0.83}& \textbf{1.98}\\
faucet& 4.46 & 4.39 & 7.2 & 3.21 & 3.48 & 7.52 & 3.61 & 3.59 & 7.25 & 5.68 & 6.79 & 14.29 & 1.81 & 2.32 & 4.91 & 1.04 & 1.83 & 3.83& \textbf{0.47}& \textbf{0.93}& \textbf{2.14}\\
filecabinet& 2.59 & 2.48 & 3.76 & 2.02 & 1.97 & 4.14 & 2.41 & 2.12 & 4.12 & 3.72 & 3.57 & 7.13 & 1.46 & 1.71 & 2.89 & 0.78 & 1.06 & 1.83& \textbf{0.65}& \textbf{0.78}& \textbf{1.29}\\
guitar& 0.65 & 0.6 & 1.25 & 0.42 & 0.38 & 1.23 & 0.57 & 0.47 & 1.42 & 0.74 & 0.89 & 5.41 & 0.44 & 0.48 & 0.76 & 0.18 & 0.27 & 0.47&\textbf{0.11}& \textbf{0.17}& \textbf{0.28}\\
helmet& 5.39 & 5.37 & 7.96 & 3.76 & 4.18 & 7.53 & 4.36 & 4.55 & 7.73 & 9.55 & 8.41 & 15.44 & 2.33 & 3.18 & 6.03 & 1.37 & 2.40 & 4.68&\textbf{0.76}& \textbf{1.24}& \textbf{3.03}\\
jar& 3.65 & 3.87 & 6.51 & 2.57 & 2.82 & 6.0 & 3.03 & 3.17 & 7.03 & 5.44 & 5.56 & 11.87 & 1.72 & 2.37 & 4.37 & 0.97 & 1.52 & 3.03& \textbf{0.65}& \textbf{0.98}& \textbf{2.01}\\
knife& 1.29 & 0.87 & 1.21 & 0.94 & 0.62 & 1.37 & 0.84 & 0.68 & 1.44 & 2.11 & 1.53 & 3.89 & 0.72 & 0.66 & 0.96 & 0.23 & 0.36 & 0.62& \textbf{0.13}& \textbf{0.23}& \textbf{0.42}\\
lamp& 3.93 & 4.23 & 6.87 & 3.1 & 3.45 & 7.02 & 3.03 & 3.39 & 8.15 & 6.82 & 7.61 & 14.22 & 1.68 & 2.43 & 5.17 & 0.88 & 1.70 & 3.88& \textbf{0.41}& \textbf{0.94}& \textbf{2.31}\\
laptop& 1.02 & 1.04 & 1.96 & 0.75 & 0.79 & 1.59 & 0.8 & 0.85 & 1.66 & 1.04 & 1.21 & 2.46 & 0.83 & 0.87 & 1.28 & 0.39 & 0.49 & 0.86&\textbf{0.33}& \textbf{0.34}& \textbf{0.44}\\
loudspeaker& 3.21 & 3.15 & 4.55 & 2.5 & 2.45 & 5.08 & 3.1 & 2.76 & 5.32 & 4.32 & 4.19 & 7.6 & 1.75 & 2.08 & 3.45 & 0.90 & 1.34 & 2.32&  \textbf{0.69}& \textbf{0.93}& \textbf{1.60}\\
mailbox& 2.44 & 2.61 & 4.98 & 1.66 & 1.74 & 5.18 & 2.16 & 2.1 & 5.1 & 3.82 & 4.2 & 10.51 & 1.15 & 1.59 & 3.42 & 0.49 & 0.91 & 2.45&\textbf{0.28}& \textbf{0.52}& \textbf{1.62}\\
microphone& 4.42 & 5.06 & 7.04 & 3.44 & 3.9 & 8.52 & 2.83 & 3.49 & 6.87 & 6.58 & 7.56 & 16.74 & 2.09 & 2.76 & 5.7 & 1.52 & 2.66 & 5.51& \textbf{0.45}& \textbf{1.24}& \textbf{2.95}\\
microwaves& 2.67 & 2.48 & 4.43 & 2.2 & 2.01 & 4.65 & 2.65 & 2.15 & 5.07 & 4.63 & 3.94 & 6.52 & 1.51 & 1.72 & 2.76 & 0.73 & 0.96 & 1.68& \textbf{0.68}& \textbf{0.77}& \textbf{1.28}\\
motorbike& 2.63 & 2.55 & 3.52 & 2.03 & 2.01 & 3.13 & 2.29 & 2.25 & 3.54 & 2.17 & 2.48 & 5.09 & 1.38 & 1.52 & 2.26 & 0.96 & 1.20 & 1.70& \textbf{0.64}& \textbf{0.89}& \textbf{1.33}\\
mug& 3.66 & 3.67 & 5.7 & 2.45 & 2.48 & 5.17 & 2.89 & 2.56 & 5.43 & 4.76 & 4.3 & 8.37 & 1.75 & 2.16 & 3.79 & 0.98 & 1.41& 2.73& \textbf{0.83}& \textbf{1.01}& \textbf{1.72}\\
piano& 3.86 & 4.04 & 6.04 & 2.64 & 2.74 & 4.83 & 2.99 & 2.89 & 5.64 & 4.57 & 5.26 & 9.26 & 1.53 & 1.82 & 3.21 & 0.95 & 1.34 & 2.69& \textbf{0.61}& \textbf{0.72}& \textbf{1.33}\\
pillow& 2.33 & 2.38 & 3.87 & 1.85 & 1.81 & 3.68 & 2.31 & 2.26 & 4.19 & 4.21 & 3.82 & 7.89 & 1.42 & 1.67 & 3.04 & 0.64 & 0.94 & 1.73&  \textbf{0.44}& \textbf{0.53}& \textbf{0.89}\\
pistol& 1.92 & 1.62 & 2.52 & 1.25 & 1.17 & 2.65 & 1.5 & 1.3 & 2.62 & 2.27 & 2.09 & 7.2 & 1.11 & 1.06 & 1.76 & 0.57 & 0.78 & 1.23& \textbf{0.36}& \textbf{0.51}& \textbf{0.81}\\
flowerpot& 4.53 & 4.68 & 6.46 & 3.32 & 3.39 & 6.04 & 3.61 & 3.45 & 6.28 & 4.83 & 5.51 & 10.68 & 2.02 & 2.48 & 4.19 & 1.32 & 1.80 & 3.10&  \textbf{0.89}& \textbf{1.16}& \textbf{2.01}\\
printer& 3.66 & 4.01 & 5.34 & 2.9 & 3.19 & 5.84 & 3.04 & 3.19 &  5.84& 5.56 & 6.06 & 9.29 & 1.56 & 2.38 & 4.24 & 0.83 & 1.63 & 2.79& \textbf{0.62}& \textbf{0.93}& \textbf{1.83}\\
remote& 1.14 & 1.2 & 1.98 & 0.99 & 0.97 & 2.04 & 1.14 & 1.17 & 2.16 & 1.74 & 2.37 & 4.61 & 0.89 & 1.05 & 1.29 & 0.39 & 0.56 & 0.77& \textbf{0.29}& \textbf{0.40}& \textbf{0.48}\\
rifle& 1.27 & 1.02 & 1.37 & 0.98 & 0.8 & 1.31 & 0.98 & 0.86 & 1.46 & 1.72 & 1.45 & 3.02 & 0.83 & 0.77 & 1.16 & 0.36 & 0.50 & 0.77& \textbf{0.25}& \textbf{0.36}& \textbf{0.58}\\
rocket& 1.37 & 1.18 & 1.88 & 1.05 & 1.04 & 1.87 & 1.04 & 1.0 & 1.93 & 1.65 & 1.61 & 3.82 & 0.78 & 0.92 & 1.44 & 0.29 & 0.58 & 0.96&  \textbf{0.18}& \textbf{0.34}& \textbf{0.77}\\
skateboard& 1.58 & 1.58 & 2.07 & 1.04 & 0.94 & 1.68 & 1.08 & 1.05 & 1.84 & 1.43 & 1.6 & 3.09 & 0.82 & 0.87 & 1.24 & 0.38 & 0.53 & 0.74&\textbf{0.22}& \textbf{0.29}& \textbf{0.40}\\
sofa& 2.22 & 2.09 & 3.14 & 1.65 & 1.61 & 2.92 & 1.93 & 1.76 & 3.39 & 2.65 & 2.53 & 4.84 & 1.35 & 1.45 & 2.32 & 0.65 & 0.85 & 1.39& \textbf{0.52}& \textbf{0.58}& \textbf{0.84}\\
stove& 2.69 & 2.63 & 3.99 & 2.07 & 2.02 & 4.72 & 2.44 & 2.16 & 4.84 & 4.03 & 3.71 & 7.15 & 1.46 & 1.72 & 3.22 & 0.75 & 1.07 & 1.91& \textbf{0.62}& \textbf{0.80}& \textbf{1.28}\\
table& 2.23 & 2.15 & 3.21 & 1.56 & 1.5 & 3.36 & 1.78 & 1.65 & 3.21 & 3.03 & 3.11 & 5.74 & 1.15 & 1.33 & 2.33 & 0.57 & 0.82 & 1.55&  \textbf{0.41}& \textbf{0.51}& \textbf{0.93}\\
telephone& 1.07 & 1.06 & 1.75 & 0.8 & 0.8 & 1.67 & 1.02 & 0.95 & 1.78 & 1.3 & 1.47 & 3.37 & 0.81 & 0.89 & 1.18 & 0.38 & 0.50 & 0.71&\textbf{0.33}& \textbf{0.36}& \textbf{0.46}\\
tower& 2.46 & 2.45 & 3.91 & 1.91 & 1.97 & 4.47 & 2.15 & 2.05 & 4.51 & 3.13 & 3.54 & 9.87 & 1.26 & 1.69 & 3.06 & 0.68 & 1.08 & 2.00&\textbf{0.46}& \textbf{0.76}& \textbf{1.37}\\
train& 1.86 & 1.68 & 2.32 & 1.5 & 1.41 & 2.37 & 1.59 & 1.44 & 2.51 & 2.01 & 2.03 & 4.1 & 1.09 & 1.14 & 1.61 & 0.64 & 0.80 & 1.14& \textbf{0.44}& \textbf{0.59}& \textbf{0.90}\\
watercraft& 1.85 & 1.69 & 2.49 & 1.46 & 1.39 & 2.4 & 1.53 & 1.42 & 2.67 & 2.1 & 2.13 & 4.58 & 1.09 & 1.12 & 1.65 & 0.51 & 0.72 & 1.13&\textbf{0.34}& \textbf{0.50}& \textbf{0.78}\\
washer& 3.47 & 3.2 & 4.89 & 2.42 & 2.31 & 6.08 & 2.92 & 2.53 & 6.53 & 5.55 & 4.11 & 7.04 & 1.72 & 2.05 & 4.19 & 0.75 & 1.12 & 2.42&\textbf{0.69}& \textbf{0.86}& \textbf{1.61}\\
\midrule
mean& 2.68 & 2.66 & 4.06 & 1.96 & 1.98 & 4.09 & 2.26 & 2.17 & 4.31 & 3.84 & 3.88 & 8.03 & 1.35 & 1.63 & 2.86 & 0.70& 1.06&1.96& \textbf{0.50}& \textbf{0.69}& \textbf{1.24}\\

\bottomrule
\end{tabular}}}
\vspace{20pt}
\end{table*}

\begin{table*}
\caption{
Detailed results CD-$\ell_1$ (multiplied by 1000) and F-Score@1\% on Projected-ShapeNet-55.}
\label{Table:PShapeNet-55} 
\centering
\setlength{\tabcolsep}{5pt}{
\adjustbox{width=\linewidth}{
\begin{tabular}{@{\hskip 5pt}>{\columncolor{white}[5pt][\tabcolsep]}l | cc|cc|cc|cc|cc|c>{\columncolor{white}[\tabcolsep][5pt]}c@{\hskip 5pt}}
\toprule
\multirow{2}{*}{ }& \multicolumn{2}{c|}{PCN~\cite{PCN}}&\multicolumn{2}{c|}{TopNet~\cite{TopNet}}&\multicolumn{2}{c|}{GRNet~\cite{GRNet}}&\multicolumn{2}{c|}{SnowflakeNet~\cite{xiang2021snowflakenet}}&\multicolumn{2}{c|}{PoinTr~\cite{yu2021pointr}} &\multicolumn{2}{c}{AdaPoinTr}\\
\cmidrule{2-13} 
& CD & F-Score & CD & F-Score & CD & F-Score & CD & F-Score & CD & F-Score & CD & F-Score\\
\midrule
airplane& 9.07& 0.703& 9.85& 0.621& 8.30& 0.726& 6.35& 0.862& 6.02& 0.830& \textbf{5.18}& \textbf{0.93}\\
bag& 18.64& 0.309& 18.69& 0.235& 14.67& 0.428& 12.86& 0.535& 12.15& 0.555& \textbf{10.93}& \textbf{0.63}\\
basket& 18.03& 0.302& 18.19& 0.252& 14.60& 0.383& 13.19& 0.478& 12.00& 0.538& \textbf{10.18}& \textbf{0.64}\\
bathtub& 20.96& 0.320& 17.19& 0.292& 13.90& 0.426& 12.16& 0.543& 11.49& 0.576& \textbf{10.05}& \textbf{0.68}\\
bed& 20.54& 0.266& 20.36& 0.214& 15.32& 0.376& 13.94& 0.462& 13.48& 0.465& \textbf{12.24}& \textbf{0.55}\\
bench& 13.01& 0.531& 13.01& 0.474& 10.55& 0.605& 9.04& 0.714& 8.49& 0.739& \textbf{7.34}& \textbf{0.83}\\
birdhouse& 20.38& 0.253& 22.00& 0.165& 16.52& 0.347& 15.24& 0.428& 14.60& 0.432& \textbf{13.27}& \textbf{0.51}\\
bookshelf& 18.32& 0.313& 18.66& 0.261& 14.10& 0.416& 12.80& 0.504& 12.59& 0.508& \textbf{11.42}& \textbf{0.60}\\
bottle& 15.21& 0.430& 16.33& 0.301& 13.00& 0.488& 11.12& 0.601& 9.83& 0.680& \textbf{8.67}& \textbf{0.77}\\
bowl& 13.96& 0.448& 15.04& 0.356& 12.05& 0.514& 10.74& 0.626& 9.64& 0.692& \textbf{8.76}& \textbf{0.76}\\
bus& 12.73& 0.482& 13.11& 0.422& 11.83& 0.504& 9.99& 0.638& 9.65& 0.661& \textbf{8.76}& \textbf{0.74}\\
cabinet& 17.86& 0.302& 17.13& 0.259& 14.57& 0.366& 12.86& 0.469& 12.48& 0.490& \textbf{11.31}& \textbf{0.57}\\
camera& 22.83& 0.228& 21.96& 0.164& 15.47& 0.391& 13.68& 0.474& 13.00& 0.493& \textbf{11.57}& \textbf{0.57}\\
can& 17.12& 0.315& 16.49& 0.285& 14.01& 0.392& 11.98& 0.517& 11.09& 0.583& \textbf{9.57}& \textbf{0.70}\\
cap& 18.20& 0.326& 19.74& 0.191& 12.50& 0.520& 12.07& 0.631& 10.08& 0.706& \textbf{8.51}& \textbf{0.84}\\
car& 12.85& 0.443& 13.61& 0.377& 12.13& 0.465& 11.20& 0.526& 10.58& 0.568& \textbf{9.77}& \textbf{0.63}\\
cellphone& 12.36& 0.502& 12.56& 0.438& 11.30& 0.515& 9.18& 0.695& 8.83& 0.706& \textbf{7.61}& \textbf{0.81}\\
chair& 15.33& 0.399& 16.29& 0.316& 12.57& 0.491& 11.07& 0.591& 10.43& 0.616& \textbf{9.12}& \textbf{0.72}\\
clock& 15.85& 0.414& 15.58& 0.345& 13.01& 0.476& 11.23& 0.606& 10.63& 0.621& \textbf{9.60}& \textbf{0.70}\\
dishwasher& 19.61& 0.252& 19.10& 0.212& 15.19& 0.316& 13.29& 0.457& 13.18& 0.470& \textbf{11.94}& \textbf{0.53}\\
display& 16.40& 0.378& 16.10& 0.318& 12.74& 0.467& 11.54& 0.571& 10.78& 0.601& \textbf{9.37}& \textbf{0.70}\\
earphone& 19.78& 0.360& 19.75& 0.257& 13.42& 0.506& 14.37& 0.534& 11.97& 0.575& \textbf{10.23}& \textbf{0.67}\\
faucet& 16.98& 0.396& 17.74& 0.243& 10.81& 0.592& 9.81& 0.669& 9.10& 0.662& \textbf{7.23}& \textbf{0.83}\\
file cabinet& 18.28& 0.287& 17.88& 0.249& 15.08& 0.353& 13.07& 0.461& 12.64& 0.492& \textbf{11.55}& \textbf{0.56}\\
flowerpot& 17.59& 0.309& 17.75& 0.245& 13.73& 0.426& 12.84& 0.489& 11.80& 0.535& \textbf{10.97}& \textbf{0.61}\\
guitar& 7.96& 0.758& 9.05& 0.668& 7.56& 0.775& 6.13& 0.881& 5.74& 0.901& \textbf{4.91}& \textbf{0.95}\\
helmet& 20.56& 0.277& 20.11& 0.215& 14.63& 0.424& 13.66& 0.496& 12.39& 0.547& \textbf{11.61}& \textbf{0.62}\\
jar& 17.82& 0.339& 18.36& 0.251& 14.08& 0.442& 12.83& 0.523& 11.56& 0.604& \textbf{10.49}& \textbf{0.68}\\
keyboard& 13.69& 0.557& 11.05& 0.586& 9.71& 0.664& 8.12& 0.766& 7.61& 0.797& \textbf{6.79}& \textbf{0.85}\\
knife& 8.04& 0.759& 8.75& 0.704& 7.52& 0.778& 5.77& 0.890& 5.66& 0.873& \textbf{4.98}& \textbf{0.93}\\
lamp& 17.50& 0.394& 17.52& 0.314& 12.73& 0.555& 10.76& 0.663& 10.21& 0.679& \textbf{9.16}& \textbf{0.79}\\
laptop& 16.05& 0.429& 13.52& 0.395& 10.83& 0.533& 9.79& 0.648& 8.93& 0.689& \textbf{8.11}& \textbf{0.78}\\
loudspeaker& 21.21& 0.262& 19.72& 0.211& 16.12& 0.354& 13.95& 0.455& 13.61& 0.475& \textbf{12.60}& \textbf{0.55}\\
mailbox& 17.97& 0.350& 16.82& 0.283& 12.52& 0.507& 10.61& 0.637& 9.84& 0.654& \textbf{8.64}& \textbf{0.78}\\
microphone& 17.75& 0.413& 17.05& 0.281& 10.45& 0.630& 10.37& 0.697& 8.64& 0.691& \textbf{7.60}& \textbf{0.79}\\
microwaves& 39.54& 0.153& 26.64& 0.170& 18.35& 0.281& \textbf{16.92}& 0.398& 17.06& 0.406& 17.43& \textbf{0.49}\\
motorbike& 12.66& 0.505& 14.33& 0.364& 10.68& 0.561& 10.13& 0.605& 9.83& 0.622& \textbf{8.99}& \textbf{0.69}\\
mug& 18.12& 0.287& 20.56& 0.179& 14.60& 0.372& 14.16& 0.428& 13.10& 0.448& \textbf{11.39}& \textbf{0.54}\\
piano& 19.15& 0.291& 19.91& 0.230& 14.46& 0.414& 13.47& 0.498& 12.42& 0.520& \textbf{10.83}& \textbf{0.61}\\
pillow& 20.04& 0.291& 19.57& 0.238& 15.15& 0.421& 14.02& 0.496& 12.57& 0.563& \textbf{11.82}& \textbf{0.66}\\
pistol& 10.93& 0.587& 12.43& 0.442& 9.69& 0.635& 8.40& 0.729& 8.25& 0.733& \textbf{7.27}& \textbf{0.81}\\
printer& 26.86& 0.217& 20.74& 0.192& 16.17& 0.353& 13.75& 0.475& 13.53& 0.466& \textbf{12.46}& \textbf{0.56}\\
remote& 14.62& 0.447& 13.52& 0.436& 12.18& 0.511& 10.07& 0.702& 9.55& 0.711& \textbf{8.81}& \textbf{0.77}\\
rifle& 8.79& 0.722& 9.11& 0.673& 7.30& 0.792& 6.10& 0.873& 5.98& 0.863& \textbf{5.21}& \textbf{0.93}\\
rocket& 10.98& 0.585& 10.45& 0.610& 8.58& 0.707& 7.49& 0.793& 6.86& 0.782& \textbf{5.58}& \textbf{0.90}\\
skateboard& 9.65& 0.662& 11.01& 0.568& 8.82& 0.704& 7.39& 0.801& 7.24& 0.775& \textbf{6.16}& \textbf{0.89}\\
sofa& 17.12& 0.297& 16.93& 0.257& 14.36& 0.373& 12.59& 0.477& 12.11& 0.502& \textbf{10.89}& \textbf{0.59}\\
stove& 22.04& 0.270& 20.33& 0.231& 16.64& 0.355& 13.77& 0.463& 13.65& 0.474& \textbf{12.62}& \textbf{0.55}\\
table& 14.79& 0.469& 14.40& 0.433& 12.01& 0.537& 10.49& 0.645& 9.97& 0.663& \textbf{8.81}& \textbf{0.75}\\
telephone& 12.02& 0.517& 12.18& 0.457& 10.95& 0.532& 8.82& 0.720& 8.62& 0.720& \textbf{7.51}& \textbf{0.81}\\
tower& 15.39& 0.432& 16.94& 0.306& 13.26& 0.493& 11.96& 0.577& 11.06& 0.600& \textbf{10.12}& \textbf{0.68}\\
train& 11.93& 0.541& 12.56& 0.462& 10.64& 0.579& 9.47& 0.665& 9.22& 0.673& \textbf{8.20}& \textbf{0.76}\\
trash bin& 15.39& 0.343& 16.38& 0.270& 14.01& 0.380& 12.62& 0.466& 11.88& 0.518& \textbf{10.83}& \textbf{0.61}\\
washer& 22.39& 0.208& 21.90& 0.162& 18.50& 0.280& 15.16& 0.401& 15.21& 0.410& \textbf{14.19}& \textbf{0.51}\\
watercraft& 12.61& 0.527& 13.16& 0.439& 10.58& 0.593& 9.17& 0.692& 8.85& 0.692& \textbf{7.90}& \textbf{0.78}\\
\midrule
mean& 16.64& 0.403& 16.35& 0.337& 12.81& 0.491& 11.34& 0.594& 10.68& 0.615& \textbf{9.58}& \textbf{0.70}\\
\bottomrule
\end{tabular}}}
\vspace{20pt}
\end{table*}

\end{document}